\documentclass{article}

\usepackage[preprint]{neurips_2026}

\usepackage[utf8]{inputenc} %
\usepackage[T1]{fontenc}    %
\usepackage{hyperref}       %
\usepackage{url}            %
\usepackage{booktabs}       %
\usepackage{amsfonts}       %
\usepackage{nicefrac}       %
\usepackage{microtype}      %
\usepackage{xcolor}         %

\usepackage{amsmath}
\usepackage{amssymb}
\usepackage{mathtools}
\usepackage{amsthm}

\usepackage{xurl}

\usepackage[capitalize,noabbrev]{cleveref}

\usepackage{booktabs} 
\usepackage{amssymb} %
\usepackage{pifont}  %
\usepackage{xcolor}  %
\usepackage{adjustbox}
\usepackage{tabularx}
\usepackage{makecell}

\usepackage{subcaption}
\usepackage{placeins}  %
\usepackage{float} %

\usepackage{listings}%
\usepackage{fvextra} %

\usepackage{pdfpages}

\DefineVerbatimEnvironment{jsonblock}{Verbatim}{
	fontsize=\footnotesize,
	breaklines=true,      %
	breakanywhere=true,   %
	samepage=false        %
}

\title{{Mini Amusement Parks (MAPs):\\A Testbed for Modelling Business Decisions}}

\author{St\'ephane Aroca-Ouellette\textsuperscript{*, 1} \And Ian Berlot-Attwell\textsuperscript{*, 1, 2, 3} \And Panagiotis Lymperopoulos\textsuperscript{1} \And Abhiramon Rajasekharan\textsuperscript{1} \And Tongqi Zhu\textsuperscript{1} \And Herin Kang\textsuperscript{1} \And Kaheer Suleman\textsuperscript{1} \And Sam Pasupalak\textsuperscript{1}}

\makeatletter
\AddToHook{cmd/appendix/before}{\def\cref@section@alias{appendix}}
\makeatother

\begin{document}

\definecolor{cec1d24}{RGB}{236,29,36}
\definecolor{cffffff}{RGB}{255,255,255}

\newcommand{\cmark}{\textcolor{green!70!black}{\textbf{\checkmark}}}
\newcommand{\xmark}{\textcolor{red}{\ding{55}}}
\newcommand{\pmark}{\textcolor{orange}{$\boldsymbol{\blacktriangle}$}}

\newcommand{\envacro}{\textsc{Maps}} %
\newcommand{\envname}{Mini Amusement Parks}

\maketitle

\vspace{-1cm}
\begin{center}
	{\footnotesize \textsuperscript{*}Equal Contribution, \textsuperscript{1}Skyfall.ai, \textsuperscript{2}University of Toronto, \textsuperscript{3}Vector Institute}
\end{center}
\vspace*{0.5cm}

\begin{abstract}
Despite rapid progress in artificial intelligence, current systems struggle with the interconnected challenges that define real-world decision making. Practical domains such as business management require open-ended optimization, actively learning environment dynamics from sparse experience, planning over long horizons in stochastic settings, and reasoning over spatial information. Yet no existing human--AI benchmarks assess how well agents integrate these challenges in a grounded decision-making context. To this end, we introduce \envname\ (\envacro), an amusement-park simulator designed to evaluate an agent’s ability to model its environment, anticipate long-term consequences under uncertainty, and strategically operate a complex business. We provide expert human performance and a comprehensive evaluation of state-of-the-art agents, finding experts outperform these systems by 11.4× on easy mode and 15.3× on medium mode. Our analysis reveals persistent weaknesses in long-horizon planning, sample-efficient learning, spatial reasoning, and modelling uncertainty. By unifying these challenges within a single environment, \envacro\ offers a new foundation for benchmarking agents capable of adaptable decision making. Code: \url{https://github.com/Skyfall-Research/MAPs}
\end{abstract}

\section{Introduction}

AI systems have achieved or surpassed human performance on a wide range of tasks and exams \citep{alphago, alphacode, bar_performance, mcat_performance}. Yet these successes are consistently in environments with narrow objectives and minimal uncertainty. This differs from many tasks in human domains: for example, running a business. Here, success hinges on coordinating interdependent decisions---allocating staff, developing infrastructure, investing in R\&D, and conducting market research---while simultaneously reasoning over long horizons about noisy, partially observable customer behavior shaped by multiple, often multi-modal, interacting factors. Moreover, business owners must actively and efficiently learn how to navigate these challenges, or risk bankruptcy.

Each of these dimensions, while frequently navigated by humans, presents a major hurdle for state-of-the-art (SotA) AI systems. (1) When applied directly, large language models (LLMs) perform poorly on long-horizon planning and sequential reasoning tasks \citep{limit_planning,reseed}, often exhibiting myopic optimization for immediate concerns \citep{revealing_planning, nonmyopic}. Mitigation methods, such as tree search or iterative reasoning \citep{tree_of_thoughts}, quickly become expensive and are limited by the model's often inadequate evaluation of nodes \cite{tot_discriminator}. (2) AI systems remain inefficient learners, struggling to infer patterns from a handful of examples \citep{arcagi}. In contrast, humans actively construct and test hypotheses to accelerate learning \citep{theory_theory} a capacity largely unexplored in current AI research (3) Existing models lag behind humans in non-linguistic reasoning, such as spatial or visual relationships \citep{multimodal_reasoning,stuckmatrixprobingspatial}. (4) Lastly, LLMs are unreliable when estimating causal and probabilistic relationships \citep{llm_statistical_causal_reasoning}, particularly when disentangling complex conditional distributions from samples \citep{reasoning_uncertainty}.  

Effective performance in real-world domains, such as business management, requires integrating all of these capabilities. To identify fruitful avenues of research and track progress toward robust, real-world decision-making, we introduce \envname\ (\envacro), an amusement-park simulator that evaluates an agent’s ability to actively model its environment, anticipate long-term outcomes, understand spatial relations, and optimize under uncertainty.

Prior benchmarks with human baselines fall into three categories.
Works such as Crafter \citep{crafter} and DiscoveryWorld \citep{discoveryworld} emphasize long-term planning through hierarchical task structures but simplify entity relationships to static prerequisites, omitting the rich interdependencies of complex systems. Abstract reasoning benchmarks such as ARC-AGI and AutumnBench \citep{arcagi2, warrier2025benchmarkingworldmodellearning} probe few-shot generalization with synthetic grid-based puzzles -- revealing substantial human–AI gaps but lacking real-world semantics. Lastly, applied benchmarks evaluate agents on practical tasks, such as OSWorld \citep{osworld} for computer-based tasks and VendingBench \citep{vendingbench} for operating a vending machine. Notably, only VendingBench evaluates open-ended optimization under uncertainty -- the remaining benchmarks are deterministic or satisfaction-based tasks.

\begin{table*}[t!]
	\centering
	\small
	\caption{Comparison of benchmarks that report human performance. \xmark\ = absence \pmark\ = partial presence \cmark\ = presence.}
	\label{tab:rel_work}
	\begin{adjustbox}{max width=\textwidth}
		\begin{tabular}{|l|ccccc|ccc|}
			\hline
			& CH1 & CH2 & CH3 & CH4 & CH5 & \multicolumn{3}{c|}{Features}  \\
			\toprule 
			Environment 
			& {\shortstack{Open-Ended\\Objective}} 
			& {\shortstack{Long-Horizon\\Planning}} 
			& {\shortstack{Active WM\\Learning}}  
			& {\shortstack{Spatial\\Reasoning}} 
			& {\shortstack{Stochastic\\Transitions}}
			& {\shortstack{Online\\Leaderboard}} 
			& {\shortstack{Variable\\Difficulty}} 
			& {\shortstack{Playable\\Online}} \\
			\midrule
			ARC-AGI2       & \xmark & \xmark & \pmark & \cmark & \xmark & \cmark & \cmark & \cmark \\
			AutumnBench   & \xmark & \pmark & \cmark & \cmark & \cmark & \xmark & \xmark & \cmark \\
			Crafter        & \xmark & \cmark & \pmark & \cmark & \xmark & \xmark & \xmark & \xmark \\
			DiscoveryWorld & \xmark & \cmark & \pmark & \cmark & \xmark & \xmark & \cmark & \xmark \\
			WebShop        & \xmark & \cmark & \pmark & \cmark & \xmark & \xmark & \xmark & \xmark \\
			WebArena       & \xmark & \cmark & \pmark & \cmark & \xmark & \cmark & \xmark & \xmark \\
			OSWorld        & \xmark & \cmark & \pmark & \cmark & \xmark & \cmark & \xmark & \xmark \\
			VendingBench   & \cmark & \cmark & \pmark & \xmark & \cmark & \cmark & \xmark & \xmark \\
			\envacro\       & \cmark & \cmark & \cmark & \cmark & \cmark & \cmark & \cmark & \cmark \\
			\bottomrule
		\end{tabular}
	\end{adjustbox}
\end{table*}

\envacro\ bridges these gaps. Like DiscoveryWorld, it requires long-horizon planning but features richer, more interwoven dynamics and incorporates stochastic transitions. Like ARC-AGI2, it demands sample efficiency and spatial reasoning, yet is grounded in a realistic domain. Like VendingBench, it emphasizes an open-ended objective under uncertainty, while additionally requiring actively learning environment dynamics and spatial reasoning. By integrating these challenges, \envacro\ provides a unique, comprehensive, and scalable AI testbed for robust business-style decision-making.

Our primary contribution is the introduction of \envacro\ and a comprehensive evaluation of humans and SotA AI systems across multiple settings. These experiments reveal five core challenges:
\textbf{CH1.} SotA models lag behind expert humans on the open-ended, multi-faceted task of business-style decision-making in \envacro, achieving at most 8.80\% of the human upper bound.
\textbf{CH2.} Performance deteriorates substantially as the demands of long-horizon planning increase, underscoring persistent myopia in current models.
\textbf{CH3.} Even with access to an exploratory sandbox, models fail to learn environment dynamics---a capability we term \textsl{active world-model learning}.
\textbf{CH4.} Agents show significant difficulty with spatial reasoning.
\textbf{CH5.} Systems struggle to model and act robustly under stochastic transitions, leading to brittle behaviour under uncertainty.
These findings demonstrate that \envacro\ exposes a broad spectrum of persistent gaps in contemporary AI systems and provides actionable direction for advancing more capable decision-making agents.

\section{Related Work}

We position \envacro\ relative to benchmarks that include both an environment and human baselines across the five challenges posed by \envacro. 
\Cref{tab:rel_work} shows an overview of related benchmarks. See \cref{app:related_works} for further related works. %

\subsection{Benchmarks without Open-Ended Objectives}
Most benchmarks with human baselines evaluate binary task completion under deterministic dynamics; this differs from the graded, noisy objectives humans routinely face.

One family of works focuses on abstract tasks far removed from real-world domains: ARC-AGI2 \citep{arcagi2} and AutumnBench \citep{warrier2025benchmarkingworldmodellearning} use grid-based colouring environments to test fluid intelligence and world modelling respectively. IVRE \citep{IVRE} probes causal reasoning in a Blicket-style environment \citep{blicket}. These benchmarks reveal human–AI gaps, but minimize real-world semantics. %
Another line of work assesses long-term planning via dependency trees. In Crafter \citep{crafter} and HeroBench \citep{herobench} %
players navigate a gridworld, find resources, craft increasingly complex items, and fight monsters; DiscoveryWorld \citep{discoveryworld} requires players to use items to iteratively perform scientific experiments and build knowledge through discoveries. Finally there are benchmarks revolving around real computer-based tasks: e.g., WebShop \citep{webshop}, WebArena \citep{webarena}, and OSWorld \citep{osworld}. These evaluate long-horizon interaction with websites or operating systems but lack open-ended optimization objectives and stochastic environment dynamics.

\subsection{Benchmarks with Open-Ended Objectives}

Few environments combine human baselines with open-ended optimization. Most closely related is VendingBench \citep{vendingbench}, a vending machine simulator with stochastic customers and a long time horizon. While it reports human performance, it includes only a single participant, does not require agents to learn or adapt from interaction---a core hallmark of human intelligence \citep{how_to_grow_a_mind}---and involves no spatial reasoning. Moreover, the benchmark is already saturated, with LLMs outperforming humans. Crafting and survival-based environments evaluate long-horizon dependency-driven reasoning, and require substantial game-specific knowledge to be successful, but typically evaluate task completion \citep{minedojo,herobench}.  The Nethack Learning Environment (NLE) \citep{nethack} is an exception, instead optimizing the game's multi-faceted score. It does not, however, map to real world tasks, nor provide an evaluation framework to support sample-efficient world modelling. Lastly, Alchemy \citep{alchemy} and ScienceWorld \cite{scienceworld} evaluate agent's ability to apply scientific principles. While these environments resemble real human problem-solving tasks, they are fully deterministic and do not require robust reasoning under uncertainty.

\section{Method}\label{sec:method}

In \textbf{\envname\ (\envacro)}, a player takes the role of an amusement park manager and must maximize the value of the park within a given time horizon. Every morning they perform an action such as building new a ride or shop, hiring staff, or setting a research agenda. Guests then spend the day interacting with the park. A visualization is shown in \cref{app:gui}.

\envacro\ consists of six main components---terrain, rides, shops, staff, subclasses and research, and guests---and can be played in easy or medium mode. We briefly outline these components below, and refer the reader to the game's official documentation for more detail  (see \cref{app:documentation}).  

The park is a 20×20 grid with an entrance, an exit, a connecting path, and water tiles that increase the excitement of nearby rides. Rides (carousels, ferris wheels, or roller coasters), drive the park's capacity and rating, which in turn determine guest attendance. Shops cater to the needs of guests. These include food and drink shops, as well as specialty shops that provide a range of facilities such as souvenir shops, ATMs, and info booths. As in a real business, inventory must be managed: over-stocking leads to waste and under-stocking causes them to go out of service. The park can employ three types of staff: janitors clean dirty tiles, mechanics repair broken rides, and specialists perform a variety tasks like restocking shops or entertaining guests.

Each ride, shop, and staff subtype has tiered subclasses: yellow, blue, green, and red. Yellow entities are basic and affordable, while red entities provide greater value at higher costs. In easy mode, everything is unlocked from the start. In medium mode, they are unlocked in order via research. 

Guests visit the park with a varying initial money, energy, hunger, thirst, and happiness. Throughout the day, guests sample what to do based on their status. While guests act according to general patterns (e.g., going to a food shop when hungry) their specific decisions (e.g., which food shop to go to) are sampled from a distribution.

\envacro' design causes complex dynamics to emerge from recent actions, and the park's overall design.

\subsection{Challenges and Research Questions}

\textbf{CH1. Open-Ended Objective}
\envacro\ is designed to highlight gaps between current AI and human capabilities. Thus,

\textsl{RQ1: How do frontier models compare to humans in the integrated setting of \envacro, which jointly tests long-horizon planning, active world-model learning, spatial reasoning, and reasoning under stochasticity?}

To answer RQ1, we implement a standard ReAct \citep{react} baseline in which we condition agents on the history of past actions and observations before thinking and generating the next action. We limit the history to the past five states and actions to avoid issues with context given the large and growing state space of the game. We include the objective of maximizing park value, the game manual, and the current settings in the system prompt. 

We additionally implement and evaluate a standard reinforcement learning (RL) pipeline using PPO \citep{ppo}. Observations are constructed by mapping the structured JSON game state into a grid-based representation of the park, along with additional vectors encoding aggregated park statistics. We experiment with two variants of our RL agent. The first, \textsl{PPO$_{\text{Full}}$}, is provided with an observation and action space identical to that of the ReAct agent. The second, \textsl{PPO$_{\text{Simple}}$}, prunes the observation space, including only the aggregated park data. It also reduces the action space to only require the action type, subtype, and subclass parameters. The remaining parameters—$(x, y)$, price, and order quantity—are filled in using baseline heuristics.

For additional details, including prompts and implementation specifics, see Appendices~\ref{app:implementation_details} and \ref{app:basepromptteplates}.

\textbf{CH2. Long-Horizon Planning}
Long-horizon planning ability is a critical for enterprise AI. To assess this ability, \envacro\ provides two difficulty modes, easy and medium, where the importance of long-horizon planning is the key differentiator. On medium difficulty the horizon is extended from 50 to 100 and research introduced. The research mechanics places heavy emphasis on planning, as players must anticipate the needs of their long-term strategy. For example: an expensive ride researched too early will be unaffordable, researching attractions that are not aligned with the current strategy is wasteful, and building a high guest-count park without researching better staff can cause cascading deterioration. Success in medium mode therefore depends on forecasting future constraints and opportunities, and sequencing investments so that necessary attractions and staff are available when needed. This design directly tests whether agents can reason across extended horizons, balance short-term gains against long-term outcomes, and plan under uncertainty.

\textsl{RQ2: How does agent performance change as long-horizon planning demands increase from easy to medium mode?}

\textbf{CH3. Active WM learning}
Humans can efficiently explore and experiment in novel environments, rapidly inferring underlying dynamics by adapting prior knowledge \citep{theory_theory,how_to_grow_a_mind,self_directed}. This capacity for targeted, uncertainty-driven exploration is central to success in business domains such as risk management, market analysis, and product development, where data collection opportunities are scarce.

Inspired by this, \envacro\ tests active world-model learning, also known as world-model induction \citep{ying2025assessingadaptiveworldmodels}. In standard active learning, systems improve task performance by selectively querying an oracle. In comparison, active world-model learning agents must design and execute experiments by interacting with the environment. Agents must recognize their own uncertainty, identify the aspects of the environment most relevant to their goal, and strategically explore to reduce that uncertainty under a limited budget. To succeed, systems must construct appropriate experimental conditions to tolerate the noisy, partially observable worlds. Notably, the goal is not to replicate the transition function, but to create substantiated and impactful theories of the environment capable of supporting long-horizon predictions over key desiderata to effectively guide a policy.

To support such exploration, \envacro\ includes a novel sandbox mode that expands the action space with auxiliary actions, such as granting funds, unlocking research, switching training layouts, and undoing actions. These enable counterfactual testing and efficiently observing the distribution of possible outcomes for a given action. The sandbox mode is designed to be used with a tight sample budget, mirroring businesses that cannot afford endless focus groups or pilot projects. The demand for sample efficiency, combined with the actions it provides for active, sample-efficient world-model learning, distinguishes the task from standard RL where millions of interactions are feasible and where exploration is not traditionally done via curated experiments. In realistic domains, agents must achieve useful world models from only tens of trials. To our knowledge, the use of exploration actions is unique in the world-modelling literature; prior works  \citep{PMID:33229515, warrier2025benchmarkingworldmodellearning} operate in simpler environments and only provide a reset action.

\textsl{RQ3: Can current LLMs effectively leverage a sandbox mode to improve their performance?}

To answer this question, we modify our ReAct agent to learn from 100 in-game days in the sandbox mode.
First, we adjust the system prompt to explain that the agent is learning in sandbox mode and should discover useful information in preparation for the final evaluation. Second, the LLM is instructed to also generate useful notes for its future self; it is informed that it will have access to these notes in both sandbox and evaluation stages.
Once the sandbox budget is depleted the notes are extracted and added to the prompt (see Appendix~\ref{app:basepromptteplates}). We repeat this experiment for a subset of models where we withhold the manual.

\textbf{CH4. Spatial Reasoning.}
Spatial reasoning plays a central role in business decision-making: storefront location, product placement on shelves, advertising layout, and logistics all hinge on understanding spatial interdependencies. Similarly, in \envacro, profitability depends critically on spatial design choices within the park.
Guest behaviour is shaped by spatial relationships: they dislike walking long distances, rides near water are more appealing, 
and specialty shops---such as ATMs or info booths---are not directly targeted by guests so must be placed in high-traffic areas to attract customers. Effective park design therefore requires reasoning about crowd flow, adjacency effects, and multi-entity spatial dependencies that evolve over time.

\textsl{RQ4: Can LLMs and VLMs exploit spatial structure in \envacro\ to outperform a simple spatial heuristic, and does access to visual input improve this capability?}

Motivated by humans' use of vision in spatial reasoning, we first augmented the ReAct prompt with an image rendering of the current park and evaluate this vision-language-model (VLM)–based approach using three image-compatible models: Qwen3-VL, Claude Sonnet-4.5, and GPT-5. To provide a simple spatial baseline, we develop a heuristic that selects attraction placements using three rules: (1) to encourage density, it sub-selects the four empty tiles closest to existing buildings and the entrance, (2) it upweights ride and downweights shop locations based on the number of adjacent water tiles (3) for specialty shops, it upweights locations based on the number of adjacent path tiles. This heuristic captures only first-order spatial dependencies between attractions and guest flow. Optimal gameplay, however, requires a richer and more adaptive understanding of multi-entity interactions, crowd dynamics, and evolving park structure.

To answer RQ4, we compare (i) a base text-only ReAct LLM, (ii) an image-conditioned ReAct VLM, and (iii) a ReAct LLM in which all $(x, y)$ placement parameters are replaced with positions generated by the heuristic. We conduct all evaluations under the easy difficulty setting to focus the comparison on spatial reasoning rather than confounding factors such as long-horizon planning.

\textbf{CH5. Stochasticity.}
Real-world decision-making is inherently stochastic: identical actions produce varying outcomes due to epistemic and aleatoric uncertainty. Like in real businesses, action outcomes in \envacro\ are non-deterministic due to probabilistic guest behaviour. Guest decisions depend on unobservable latent variables (e.g., hunger and thirst), partially controllable factors (e.g., proximity, cleanliness, the availability of unique attractions), and irreducible randomness arising from individual sampling. Understanding this variability is essential: agents must distinguish genuine performance trends from random fluctuations, avoiding overreaction to noise while still learning from the feedback.

This is increasingly important when combined with long-horizon planning. World models (WMs) are a promising approach for planning, yet many recent works are limited to deterministic environments \citep{worldcoder, DBLP:conf/nips/DaineseMAM24}, evaluated under deterministic assumptions \citep{poe}, 
and/or cannot scale to the complexity of \envacro\ \citep{programInductionForPOMDBEstimation, khan2025lifelearninferringsymbolic}. Among scalable approaches, the most applicable prompt LLMs for state prediction \citep{hao-etal-2023-reasoning, DBLP:conf/nips/ZhaoLH23}. WALL-E is one such model, augmenting an LLM with learned rules \citep{walle}.

\textsl{RQ5: How much variability exists in \envacro, and can current AI systems model and act effectively under this uncertainty?}

To quantify stochasticity, we measure both per-step and trajectory-level variance across repeated runs. We replay identical trajectories ten times using the same sequence of actions and compute the coefficient of variation, $CV=\frac{\sigma}{\mu}$, for three key metrics: revenue, money, and park value. Revenue variance captures the unpredictability of guest behaviour. Money variance reflects how stochasticity propagates to resource constraints for actions. Park value variance assesses \envacro's consistency in evaluating performance. For per-step variance we repeatedly set the environment and execute the action. For trajectory-level variance, we repeat entire playthroughs using identical action sequences and discarding runs that diverge due to invalid actions.

Finally, to evaluate how effectively AI systems can model and act under uncertainty, we extend the ReAct baseline with model-predictive control (MPC) using either (i) an oracle WM or (ii) WALL-E as a learned WM \citep{walle}. In both we adopt random-shooting MPC—performing five rollouts under the LLM policy and selecting the first action in the best trajectory after four WM-simulated steps.

\subsection{Observation \& Action Space}
To support both human and AI play, \envacro\ provides two equivalent observation modes: a \emph{graphical interface} and a \emph{textual JSON representation} (via Pydantic). Both contain identical information, including aggregate statistics (e.g., profit, park rating, guest count) and detailed states of employees and attractions. The GUI is shown in \cref{app:gui}, and an example JSON observation in \cref{app:obs_example}. Actions are strings formatted as \texttt{Python} function calls. Available actions include \textsl{place}, \textsl{move},  and \textsl{remove} attractions or staff; \textsl{modify} attractions; \textsl{set research}; \textsl{wait} (no-op); and \textsl{survey guests}. Full action specifications are in the documentation.

\subsection{Evaluation Protocols}
Beyond difficulty settings, we define four resource settings that vary the amount of prior knowledge and experience provided to an agent:
(1) Documentation: only the manual. (2) Few-shot: only $100$ in-game days in the sandbox mode (excluding sandbox-specific exploration actions). (3) Few-shot$+$documentation: $100$ days in the sandbox mode, plus the manual. (4) Unlimited: no restrictions. 

Unless otherwise noted, agents are evaluated in the documentation only setting.  Evaluation is on three held-out layouts (i.e., not provided in sandbox training), with three random seeds per layout. The primary metric is the park's final value: the sum of final cash on hand, the sell price of rides and shops, and a small contribution from intellectual property accrued through research.

To establish an aspirational ceiling, we collect human performance through an online leaderboard. %
As participants can retry layouts and share strategies, scores reflect the unlimited resource setting---the best performance achievable with full information and unrestricted practice. This mirrors the convention of reporting \textit{expert} human performance in game benchmarks \citep{ale}, where the goal is to define how far AI must travel rather than to match a na\"{i}ve human baseline. These scores therefore serve as an empirical upper bound, not a within-condition comparison to the constrained settings in which AI agents are evaluated.
To facilitate aggregation, scores are normalized by this human ceiling for each layout and difficulty, with 100\% indicating parity.

\begin{table}[t]
	\small
	\begin{minipage}[t]{0.55\linewidth}
		\centering
		\caption{Model performance on \envacro. Human score is the mean park value under the unlimited resource setting (see \cref{sec:method}). Score (\%) is performance relative to this human ceiling, with 100.00\% indicating parity.}
		\label{tab:main_results}
		\setlength{\tabcolsep}{3pt}
		\begin{tabular}{llll}
			\toprule
			Model & Score (\%) & Runtime (s) & Cost (USD) \\
			\midrule
			\multicolumn{4}{c}{\textbf{EASY DIFFICULTY}} \\
			GPT-5 Nano & 0.29$_{\pm 0.29}$ & 1,311$_{\pm 305}$ & 0.13$_{\pm0.02}$ \\
			GPT-5 & 8.80$_{\pm 8.49}$ & 3,774$_{\pm 505}$ & 3.75$_{\pm0.32}$ \\
			Grok 4 & 2.82$_{\pm 4.50}$ & 2,908$_{\pm 437}$ & 6.60$_{\pm0.63}$ \\
			Sonnet 4.5 & 2.28$_{\pm 2.02}$ & 937$_{\pm 124}$ & 6.06$_{\pm0.60}$ \\
			Gemini 2.5 Pro & 1.82$_{\pm 2.78}$ & 787$_{\pm 84}$ & 3.06$_{\pm0.30}$ \\
			Qwen3 VL & 0.36$_{\pm 0.29}$ & 529$_{\pm 120}$ & 0.38$_{\pm0.04}$ \\
			PPO$_{\text{Full}}$ & 0.50$_{\pm 0.49}$ & 0$_{\pm 0}$ & 0.00$_{\pm0.00}$ \\
			PPO$_{\text{Simple}}$ & 7.83$_{\pm 4.33}$ & 0$_{\pm 0}$ & 0.00$_{\pm0.00}$ \\
			\multicolumn{4}{c}{\textbf{MEDIUM DIFFICULTY}} \\
			GPT-5 Nano & 0.44$_{\pm 0.47}$ & 2,506$_{\pm 705}$ & 0.27$_{\pm0.04}$ \\
			GPT-5 & 4.44$_{\pm 4.01}$ & 7,583$_{\pm 764}$ & 8.36$_{\pm0.48}$ \\
			Grok 4 & 0.44$_{\pm 0.16}$ & 5,438$_{\pm 683}$ & 13.00$_{\pm0.86}$ \\
			Sonnet 4.5 & 1.06$_{\pm 0.80}$ & 1,713$_{\pm 86}$ & 12.19$_{\pm1.03}$ \\
			Gemini 2.5 Pro & 0.52$_{\pm 0.22}$ & 1,423$_{\pm 126}$ & 5.97$_{\pm0.45}$ \\
			Qwen3 VL & 0.27$_{\pm 0.21}$ & 1,252$_{\pm 329}$ & 0.77$_{\pm0.06}$ \\
			PPO$_{\text{Full}}$ & 0.25$_{\pm 0.23}$ & 0$_{\pm 0}$ & 0.00$_{\pm0.00}$ \\
			PPO$_{\text{Simple}}$ & 2.08$_{\pm 1.76}$ & 0$_{\pm 0}$ & 0.00$_{\pm0.00}$ \\
			\bottomrule
		\end{tabular}
	\end{minipage}
	\hfill
	\begin{minipage}[t]{0.43\linewidth}
		\centering
		\caption{Scores when models are provided with and without the opportunity to learn in sandbox for 100 in-game steps. Score (\%) indicates the score relative to the human upper bound (unlimited resource setting).}\label{tab:main_sandbox}
		\setlength{\tabcolsep}{4pt}
		\begin{tabular}{lll}
			\toprule
			LLM & \makecell{Score (\%)\\(No Learning)} & \makecell{Score (\%)\\(Learning)} \\
			\midrule
			\multicolumn{3}{c}{\textbf{EASY DIFFICULTY}} \\
			GPT-5 Nano & 0.29$_{\pm 0.29}$ & 0.10$_{\pm 0.09}$ \\
			GPT-5 & 8.80$_{\pm 8.49}$ & 8.70$_{\pm 7.53}$ \\
			Grok 4 & 2.82$_{\pm 4.50}$ & 1.37$_{\pm 0.89}$ \\
			Sonnet 4.5 & 2.28$_{\pm 2.02}$ & 0.44$_{\pm 0.42}$ \\
			Gemini 2.5 Pro & 1.82$_{\pm 2.78}$ & 0.63$_{\pm 0.58}$ \\
			\multicolumn{3}{c}{\textbf{MEDIUM DIFFICULTY}} \\
			GPT-5 Nano & 0.44$_{\pm 0.47}$ & 0.09$_{\pm 0.08}$ \\
			GPT-5 & 4.44$_{\pm 4.01}$ & 6.52$_{\pm 4.81}$ \\
			Grok 4 & 0.44$_{\pm 0.16}$ & 0.85$_{\pm 0.57}$ \\
			Sonnet 4.5 & 1.06$_{\pm 0.80}$ & 0.70$_{\pm 0.48}$ \\
			Gemini 2.5 Pro & 0.52$_{\pm 0.22}$ & 0.42$_{\pm 0.45}$ \\
			\bottomrule
		\end{tabular}
	\end{minipage}
\end{table}

\section{Results}
We organize our quantitative and qualitative findings by challenge. For further results %
see \cref{app:detailedresults}.

\subsection{CH1. Open-Ended Objective}\label{sec:open_end}
We benchmark GPT-5 Nano, GPT-5 \cite{gpt5}, 
Grok-4 \cite{grok4}, Claude Sonnet 4.5 \cite{claude_sonnet}, Gemini 2.5 Pro \cite{gemini2_5}, Qwen3-VL-235B-Instruct \cite{qwen3}, and the two PPO-based RL agents \cite{ppo} on \envacro. \cref{tab:main_results} shows that all benchmarked systems substantially lag experts; even the best-performing model, GPT-5, achieves an average park value that is only 8.80\% of the human upper bound. %
Moreover, GPT-5 requires over 2 hours to complete a game, compared to $\sim$ 50 minutes for humans.

Comparing the PPO variants, agents trained with a simplified observation and action space \textsl{PPO$_\text{Simple}$} outperforms agents operating over the full observation and action space \textsl{PPO$_\text{Full}$}. We hypothesize that this gap arises from credit assignment and action sensitivity. Although the heuristics used to complete missing parameters in \textsl{PPO$_\text{Simple}$} are suboptimal, they allow the agent to more reliably associate high-level action choices with downstream rewards. In contrast, in \textsl{PPO$_\text{Full}$}, a reasonable action, such as selecting an appropriate ride type, can be rendered ineffective by a single poorly chosen parameter (e.g., a ticket price of zero), substantially exacerbating the credit assignment problem. This challenge is compounded by the large observation space, which makes it difficult to identify the most relevant markers for the decision at hand. Note, the \textsl{PPO$_\text{Full}$} setting mirrors the observation and action interface used by the ReAct-based LLM agents. The relative strength of LLMs in this regime suggests that they can leverage their prior knowledge to mitigate some of these challenges.

Reviewing medium difficulty ReAct trajectories, we identified three consistent failure patterns:
(1) Sub-optimal research strategies. GPT-5 uses research most effectively, a key reason for its higher scores; we discuss this further in \cref{sec:long_horizon}. (2) Poor spatial placement. Parks tend to lack density, miss water adjacency, and cluster identical ride types; we discuss this further in \cref{sec:spatial}. (3) Excessive waiting: Claude and Gemini wait in over 40\% of turns, whereas GPT-5 waits in only 2.65\%.
Further qualitative observations, including under-utilisation of shops, lack of ride diversity, and model-specific behaviours, can be found in \cref{app:qual_succ}.

\subsection{CH2: Long-Horizon Reasoning} %
\label{sec:long_horizon}

\begin{table}[t]
\small
\begin{minipage}[t]{0.48\linewidth}
	\centering
	\caption{The impact of long-horizon planning is seen in the relative performance drop from easy to medium mode. Medium difficulty doubles the horizon and emphasizes long-term planning by introducing  research as a game mechanic. Score (\%) indicates the score relative to the human upper bound (unlimited resource setting), with 100.00\% indicating parity.}
	\label{tab:long_hor}
	\setlength{\tabcolsep}{3pt}
	\begin{tabular}{lll}
		\toprule
		LLM & Easy Score (\%) & Med. Score (\%) \\
		\midrule
		GPT-5 Nano & 0.29$_{\pm 0.29}$ & 0.44$_{\pm 0.47}$ \\
		GPT-5 & 8.80$_{\pm 8.49}$ & 4.44$_{\pm 4.01}$ \\
		Grok 4 & 2.82$_{\pm 4.50}$ & 0.44$_{\pm 0.16}$ \\
		Sonnet 4.5 & 2.28$_{\pm 2.02}$ & 1.06$_{\pm 0.80}$ \\
		Gemini 2.5 Pro & 1.82$_{\pm 2.78}$ & 0.52$_{\pm 0.22}$ \\
		Qwen3 VL & 0.36$_{\pm 0.29}$ & 0.27$_{\pm 0.21}$ \\
		\bottomrule
	\end{tabular}
\end{minipage}
\hfill
\begin{minipage}[t]{0.50\linewidth}
	\centering
	\caption{Easy mode performance when providing the model with text-only, text and the current MAPs frame, and with a simple placement heuristic that overrides the LLMs choice of coordinates. The spatial reasoning of LLMs currently underperforms a simple heuristic. Score (\%) indicates the score relative to the human upper bound (unlimited resource setting), with 100.00\% indicating parity.}\label{tab:spatial}
	\setlength{\tabcolsep}{3pt}
	\begin{tabular}{llll}
		\toprule
		LLM & Text (\%) & \makecell{Vision\\+Text (\%)} & Heuristic (\%) \\
		\midrule
		GPT-5 & 8.80$_{\pm 8.49}$ & 5.72$_{\pm 9.85}$ & 14.94$_{\pm 13.44}$ \\
		Sonnet 4.5 & 2.28$_{\pm 2.02}$ & 1.28$_{\pm 1.75}$ & 5.03$_{\pm 4.25}$ \\
		Qwen3 VL & 0.36$_{\pm 0.29}$  & 0.15$_{\pm 0.16}$ & 0.52$_{\pm 0.32}$ \\
		\bottomrule
	\end{tabular}
\end{minipage}
\end{table}

To evaluate the impact of long-horizon planning, we compare the relative performance of models under easy and medium difficulty. I.e., amplifying the need for foresight and strategic sequencing by extending the planning horizon and introducing research. As shown in \cref{tab:long_hor}, relative performance to humans declines significantly. GPT-5, the most performant model we tested, drops from 8.80\% to 4.44\%, indicating that even the strongest systems struggle with increased temporal and structural complexity. Notably, human players achieve higher park values in medium than easy, as the longer-horizon outweighs the additional challenge of research; the inverse holds for AI systems. 

Examining the research patterns of models reveals several consistent trends. All models exhibit a strong bias toward rides, with roller coasters being the most frequent. This choice is typically suboptimal, as the blue roller coaster is prohibitively expensive. Excluding GPT-5, in 13 of the 14 trajectories where it is researched, it is never built. While GPT-5 over prioritizes blue roller coasters, it is the only model to consistently use all of its unlocked research, placing at least one instance of every researched entity. Moreover, GPT-5 with sandbox learning is the only agent to reach the green tier, completing a single run with both a green roller coaster and a green ferris wheel. A possible reason for this success is that GPT-5 continues expansion until profits can sustain research; Grok-4 and Gemini create small parks and wait until they can afford a single research step, leading to inferior performance. Only three agents---GPT-5 (with learning) and Grok-4 (with and without learning)---research shops, and this occurs in fewer than one-third of their trajectories. Staff research virtually absent, with a single observed case of GPT-5 with learning unlocking blue janitors and mechanics. This diverges sharply from expert human play, where every entity type is typically researched to at least the blue level (further progression depends on park layout and strategic intent).

Overall, these findings suggest that current LLM-based agents adopt a myopic approach to success. Rather than planning for future park requirements and constraints, they pursue greedy strategies that fail to account for the long-term dependencies imposed by research and expansion.

\subsection{CH3: Active World Modelling}
\label{sec:spatial}

\cref{tab:main_sandbox} shows the results of the tested models with no learning and with the learnings from interacting in the sandbox mode for 100 in-game days. We additionally evaluate GPT-5 Nano and GPT-5 both with and without learning when the game's documentation is not provided; see \cref{tab:learning}.

In most cases, the average performance with learning is unchanged or \textit{decreases}. Based on qualitative analysis, we hypothesize this is due to the learnings generated: we find they often parrot the same concepts provided in the manual, are overly specific to the current transition, fail to consider stochasticity or necessary context in the state, or fail to capture true causal relationships by attributing changes to their current action rather than the overall park design. The few correct insights are not clearly actionable, or are for late stages of the game that the model never reaches in evaluation. Overall, learnings provide little to no benefit, merely diluting the context.

There are two exceptions: in medium mode, Grok-4 improves from 0.44 to 0.85\% and GPT-5 improves from from 4.44 to 6.52\%. The issues are lessened, and a few general and more actionable insights are identified. With learning GPT-5 becomes more successful at research, though whether this is due to a more directed approach, or simply a more successful early game that enables research faster is unclear. A breakdown of a subset of learnings for each model can be found in \cref{app:learnings}.

Without the game documentation, we see an expected drop in GPT-5 performance. Qualitatively, GPT-5 acts more reactively, impatiently selling off attractions over repositioning or adjusting inventory. Interestingly, without the documentation, sandbox learning produces an average improvement in easy mode (2.90\% $\rightarrow$ 5.95\%, \cref{tab:learning}), but drops performance in medium (3.52\% $\rightarrow$ 0.77\%, \cref{tab:learning}). The drop's root cause is unclear, but we observe an increased tendency to sell attractions -- see~\cref{app:qual_succ}

\subsection{CH4: Spatial Reasoning}
We evaluate three variants of the ReAct policy---text-only, text+image, and text with a spatial heuristic---on three image-capable models: Qwen3-VL, Claude Sonnet 4.5, and GPT-5. As shown in \cref{tab:spatial}, providing an image of the park reduces performance across all three models, suggesting that the lack of a visual modality is not the primary bottleneck for spatial reasoning in \envacro, and that the added context generates more confusion than guidance. In contrast, incorporating the spatial heuristic consistently improves performance. Heuristic-guided layouts are more compact, with rides more frequently near water, and shops more often at intersections (see \cref{app:spatial}). The heuristic also disrupts the harmful tendency of base models, particularly GPT-5, to cluster identical ride types.

We also observe failures in path reasoning: agents sometimes construct rides that are inaccessible in layouts with disconnected path, and waste turns recovering from this by selling or relocating them. Relatedly, GPT-5 occasionally clusters rides that are nearby in Euclidean distance but far apart when accounting for winding paths, further indicating an incomplete understanding of path structure.

Park designs is far from optimal, making spatial planning as a promising direction for future work.

\subsection{CH5: Stochasticity}

\begin{table}[t]
	\small
	\begin{minipage}[t]{0.43\linewidth}
		\centering
		\includegraphics[width=0.95\linewidth]{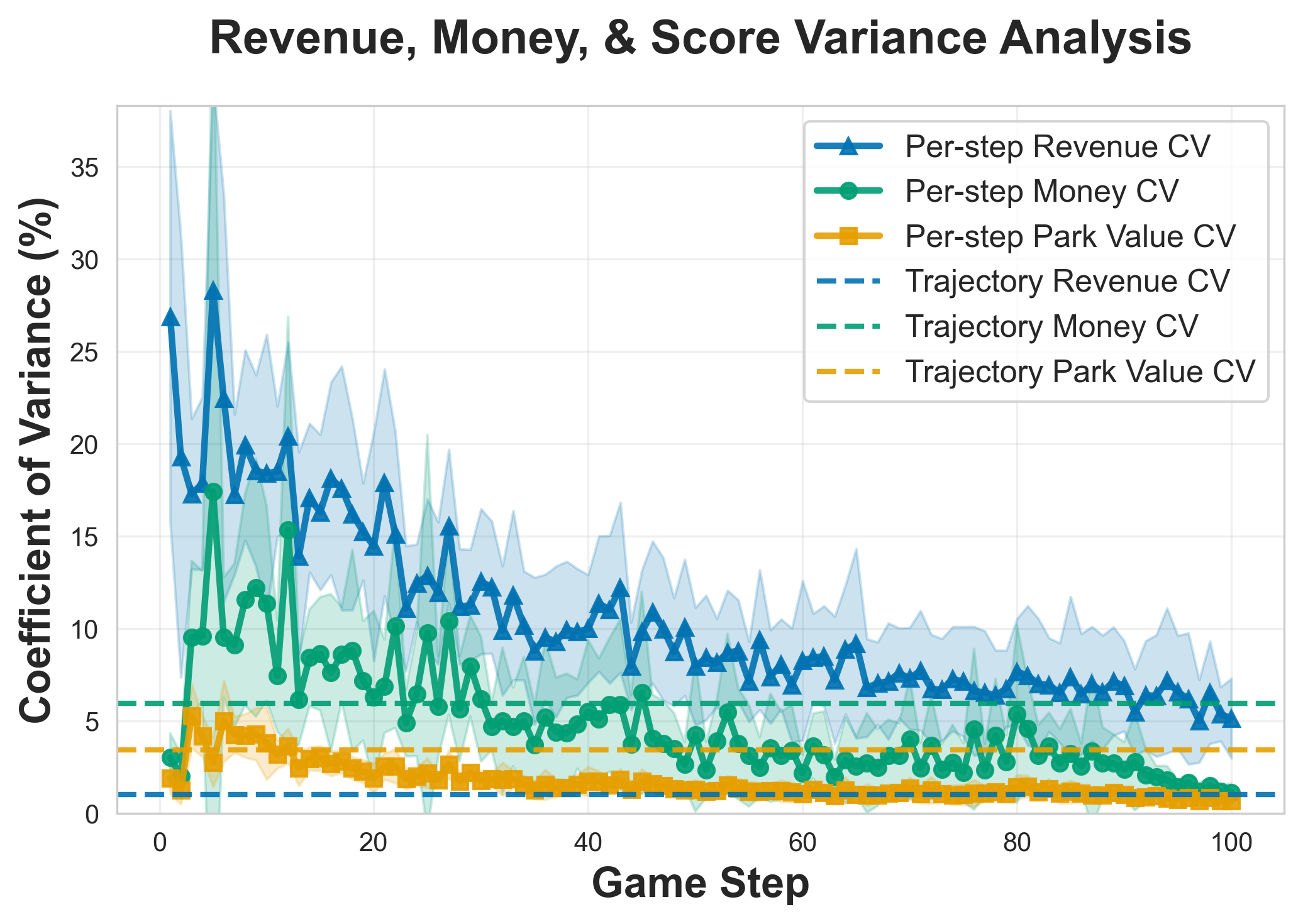}
		\captionof{figure}{The per-day and full trajectory coefficients of variation for revenue, money, and park value across several full games.}
		\label{fig:stochastic}
	\end{minipage}
	\hfill
	\begin{minipage}[b]{0.55\linewidth}
		\centering
		\caption{Planning via a World Model (WM) and Model Predictive Control (MPC) is understudied in stochastic environments such as \envacro. While the WALL-E \citep{walle} WM degrades performance, an oracle WM yields a $\sim$3.5x improvement, outperforming most frontier models tested on medium difficulty. Score (\%) is relative to the human upper bound (unlimited resource setting).}
		\label{tab:stochastic}  %
		\setlength{\tabcolsep}{3pt}
		\begin{tabular}{llll}
			\toprule
			Agent & Score (\%) & Runtime (s) & Cost (USD) \\
			\midrule
			\multicolumn{4}{c}{\textbf{EASY DIFFICULTY}} \\
			GPT-5 Nano & 0.29$_{\pm 0.29}$ & 1,311$_{\pm 305}$ & 0.13$_{\pm0.02}$ \\
			GPT-5 Nano$_\text{WALL-E}$ & 0.06$_{\pm 0.08}$ & 27,238$_{\pm 3,524}$ & 7.22$_{\pm0.61}$ \\
			GPT-5 Nano$_\text{Oracle}$ & 0.96$_{\pm 0.49}$ & 10,787$_{\pm 1,218}$ & 3.78$_{\pm0.30}$ \\
			\midrule
			\multicolumn{4}{c}{\textbf{MEDIUM DIFFICULTY}} \\
			GPT-5 Nano & 0.44$_{\pm 0.47}$ & 2,506$_{\pm 705}$ & 0.27$_{\pm0.04}$ \\
			GPT-5 Nano$_\text{Oracle}$ & 1.60$_{\pm 0.68}$ & 22,725$_{\pm 3,715}$ & 8.09$_{\pm1.31}$ \\
			\bottomrule
		\end{tabular}
	\end{minipage}
\end{table}

\cref{fig:stochastic} illustrates the variance in \envacro, plotting the coefficient of variation, per-step and per-trajectory. Note that as the game progresses and the number of guests increases, the variance decreases due to the law of large numbers. Comparing variables, revenue has the highest variance indicating that guest behaviour is non-deterministic. Money (i.e. the budget limiting possible actions) also has a relatively high-variance, showing that agents must have a robust policy that adapts to the outcomes that arise. Lastly, park value, the main evaluation metric, has a comparatively low variance, demonstrating that \envacro\ can reliably detect differences in performance. We note that order of the per-trajectory CVs is largely dictated by the inverse means. This is because the revenue's variance is passed on to money and park value, resulting in all three having a similar variance.

We next look at how current systems handle  non-determinism. \cref{tab:stochastic} highlights that directly using an LLM to predict stochastic outcomes (i.e., the Wall-E WM) is ineffective: MPC with these predictions degrades performance, is $\sim$8 times slower, and is $\sim$29 times more expensive than using the policy directly. However, an oracle WM boosts performance with a $>$3x improvement; under medium mode the oracle MPC even outperforms Sonnet-4.5 on average (1.06$_{\pm 0.8}$ $\rightarrow$ 1.60$_{\pm 0.68}$) at a lower cost (12.19$_{\pm1.03}$ $\rightarrow$ 8.09$_{\pm1.31}$), though speed remains an issue (1,713$_{\pm 86}$ $\rightarrow$ 22,725$_{\pm 3,715}$). Notably, these benefits occur with a mere 4-step look-ahead, whereas humans often plan much further ahead. %
These highlights that world modelling holds promise, but work addressing accurate stochastic modelling, inference speed, and search speed is required.

\section{Conclusion.} Real-world decision-making demands agents that can plan over long horizons, learn efficiently from limited interactions, reason about spatial and causal structure, and act robustly under uncertainty. Our results show that even the strongest contemporary AI systems fall far short of expert human performance when these capabilities must be integrated. By introducing \envacro---a scalable, realistic, and diagnostically rich benchmark---we provide a unified testbed for measuring progress on these challenges. We hope this environment, along with the human and model evaluations presented here, inspires new approaches for building agents that can reason, learn, and act with the flexibility and robustness required in complex domains.

\begin{ack}
Use unnumbered first level headings for the acknowledgments. All acknowledgments
go at the end of the paper before the list of references. Moreover, you are required to declare
funding (financial activities supporting the submitted work) and competing interests (related financial activities outside the submitted work).
More information about this disclosure can be found at: \url{https://neurips.cc/Conferences/2026/PaperInformation/FundingDisclosure}.

Do {\bf not} include this section in the anonymized submission, only in the final paper. You can use the \texttt{ack} environment provided in the style file to automatically hide this section in the anonymized submission.
\end{ack}

\bibliography{main}

\begin{thebibliography}{10}

\bibitem{akiba2019optuna}
T.~Akiba, S.~Sano, T.~Yanase, T.~Ohta, and M.~Koyama.
\newblock Optuna: A next-generation hyperparameter optimization framework.
\newblock In {\em Proceedings of the 25th ACM SIGKDD International Conference
  on Knowledge Discovery and Data Mining}, pages 2623--2631, 2019.

\bibitem{PMID:33229515}
K.~R. Allen, K.~A. Smith, and J.~B. Tenenbaum.
\newblock Rapid trial-and-error learning with simulation supports flexible tool
  use and physical reasoning.
\newblock {\em Proceedings of the National Academy of Sciences of the United
  States of America}, 117(47):29302—29310, November 2020.

\bibitem{herobench}
P.~Anokhin, R.~Khalikov, S.~Rebrikov, V.~Volkov, A.~Sorokin, and
  V.~Bissonnette.
\newblock Herobench: A benchmark for long-horizon planning and structured
  reasoning in virtual worlds, 2025.

\bibitem{claude_sonnet}
{Anthropic}.
\newblock Introducing claude sonnet 4.5.
\newblock \url{https://www.anthropic.com/news/claude-sonnet-4-5}, Sept. 2025.

\bibitem{reseed}
S.~Aroca-Ouellette, K.~von~der Wense, and A.~Roncone.
\newblock {R}e{S}eeding latent states for sequential language understanding.
\newblock In C.~Christodoulopoulos, T.~Chakraborty, C.~Rose, and V.~Peng,
  editors, {\em Proceedings of the 2025 Conference on Empirical Methods in
  Natural Language Processing}, pages 25233--25247, Suzhou, China, Nov. 2025.
  Association for Computational Linguistics.

\bibitem{vendingbench}
A.~Backlund and L.~Petersson.
\newblock Vending-bench: A benchmark for long-term coherence of autonomous
  agents, 2025.

\bibitem{stuckmatrixprobingspatial}
M.~Bai, A.~K. Cohen, E.~Koss, and C.~Lichtenbaum.
\newblock Stuck in the matrix: Probing spatial reasoning in large language
  models, 2025.

\bibitem{ale}
M.~G. {Bellemare}, Y.~{Naddaf}, J.~{Veness}, and M.~{Bowling}.
\newblock The arcade learning environment: An evaluation platform for general
  agents.
\newblock {\em Journal of Artificial Intelligence Research}, 47:253--279, jun
  2013.

\bibitem{bergstra2011algorithms}
J.~Bergstra, R.~Bardenet, Y.~Bengio, and B.~K{\'e}gl.
\newblock Algorithms for hyper-parameter optimization.
\newblock In {\em Advances in Neural Information Processing Systems},
  volume~24, 2011.

\bibitem{tot_discriminator}
Z.~Chen, M.~White, R.~Mooney, A.~Payani, Y.~Su, and H.~Sun.
\newblock When is tree search useful for {LLM} planning? it depends on the
  discriminator.
\newblock In L.-W. Ku, A.~Martins, and V.~Srikumar, editors, {\em Proceedings
  of the 62nd Annual Meeting of the Association for Computational Linguistics
  (Volume 1: Long Papers)}, pages 13659--13678, Bangkok, Thailand, Aug. 2024.
  Association for Computational Linguistics.

\bibitem{babyai}
M.~Chevalier-Boisvert, D.~Bahdanau, S.~Lahlou, L.~Willems, C.~Saharia, T.~H.
  Nguyen, and Y.~Bengio.
\newblock Baby{AI}: First steps towards grounded language learning with a human
  in the loop.
\newblock In {\em International Conference on Learning Representations}, 2019.

\bibitem{arcagi}
F.~Chollet.
\newblock On the measure of intelligence, 2019.

\bibitem{arcagi2}
F.~Chollet, M.~Knoop, G.~Kamradt, B.~Landers, and H.~Pinkard.
\newblock Arc-agi-2: A new challenge for frontier ai reasoning systems, 2025.

\bibitem{self_directed}
N.~Collignon.
\newblock {\em Self-directed learning in new and changing environments:
  understanding human algorithms for exploration}.
\newblock PhD thesis, University of Edinburgh, Jul 25 2020.
\newblock PhD Thesis, Informatics, School of Informatics.

\bibitem{programInductionForPOMDBEstimation}
A.~Curtis, H.~Tang, T.~Veloso, K.~Ellis, J.~B. Tenenbaum,
  T.~Lozano{-}P{\'{e}}rez, and L.~P. Kaelbling.
\newblock Llm-guided probabilistic program induction for {POMDP} model
  estimation.
\newblock {\em CoRR}, abs/2505.02216, 2025.

\bibitem{DBLP:conf/nips/DaineseMAM24}
N.~Dainese, M.~Merler, M.~Alakuijala, and P.~Marttinen.
\newblock Generating code world models with large language models guided by
  monte carlo tree search.
\newblock In A.~Globersons, L.~Mackey, D.~Belgrave, A.~Fan, U.~Paquet, J.~M.
  Tomczak, and C.~Zhang, editors, {\em Advances in Neural Information
  Processing Systems 38: Annual Conference on Neural Information Processing
  Systems 2024, NeurIPS 2024, Vancouver, BC, Canada, December 10 - 15, 2024},
  2024.

\bibitem{minedojo}
L.~Fan, G.~Wang, Y.~Jiang, A.~Mandlekar, Y.~Yang, H.~Zhu, A.~Tang, D.-A. Huang,
  Y.~Zhu, and A.~Anandkumar.
\newblock Minedojo: Building open-ended embodied agents with internet-scale
  knowledge.
\newblock In {\em Thirty-sixth Conference on Neural Information Processing
  Systems Datasets and Benchmarks Track}, 2022.

\bibitem{gemini2_5}
{Gemini Team, Google DeepMind}.
\newblock Gemini 2.5: Pushing the frontier with advanced reasoning,
  multimodality, long context, and next generation agentic capabilities.
\newblock Technical Report arXiv:2507.06261, Google DeepMind, July 2025.

\bibitem{theory_theory}
A.~Gopnik, A.~Meltzoff, and P.~Kuhl.
\newblock {\em The Scientist in the Crib: Minds, Brains, and How Children
  Learn}.
\newblock Harper Collins, 1999.

\bibitem{blicket}
A.~Gopnik and D.~M. Sobel.
\newblock Detecting blickets: How young children use information about novel
  causal powers in categorization and induction.
\newblock {\em Child Development}, 71(5):1205--1222, 2000.

\bibitem{crafter}
D.~Hafner.
\newblock Benchmarking the spectrum of agent capabilities.
\newblock In {\em International Conference on Learning Representations}, 2022.

\bibitem{hao-etal-2023-reasoning}
S.~Hao, Y.~Gu, H.~Ma, J.~Hong, Z.~Wang, D.~Wang, and Z.~Hu.
\newblock Reasoning with language model is planning with world model.
\newblock In H.~Bouamor, J.~Pino, and K.~Bali, editors, {\em Proceedings of the
  2023 Conference on Empirical Methods in Natural Language Processing}, pages
  8154--8173, Singapore, Dec. 2023. Association for Computational Linguistics.

\bibitem{multimodal_reasoning}
Y.~Hao, J.~Gu, H.~W. Wang, L.~Li, Z.~Yang, L.~Wang, and Y.~Cheng.
\newblock Can {MLLM}s reason in multimodality? {EMMA}: An enhanced multimodal
  reasoning benchmark.
\newblock In {\em Forty-second International Conference on Machine Learning},
  2025.

\bibitem{discoveryworld}
P.~Jansen, M.-A. C{\^o}t{\'e}, T.~Khot, E.~Bransom, B.~D. Mishra, B.~P.
  Majumder, O.~Tafjord, and P.~Clark.
\newblock Discoveryworld: A virtual environment for developing and evaluating
  automated scientific discovery agents.
\newblock In {\em The Thirty-eight Conference on Neural Information Processing
  Systems Datasets and Benchmarks Track}, 2024.

\bibitem{bar_performance}
D.~M. Katz, M.~J. Bommarito, S.~Gao, and P.~D. Arredondo.
\newblock Gpt-4 passes the bar exam.
\newblock {\em SSRN}, 2023.

\bibitem{khan2025lifelearninferringsymbolic}
Z.~Khan, A.~Prasad, E.~Stengel-Eskin, J.~Cho, and M.~Bansal.
\newblock One life to learn: Inferring symbolic world models for stochastic
  environments from unguided exploration, 2025.

\bibitem{nethack}
H.~K\"{u}ttler, N.~Nardelli, A.~Miller, R.~Raileanu, M.~Selvatici,
  E.~Grefenstette, and T.~Rockt\"{a}schel.
\newblock The nethack learning environment.
\newblock In H.~Larochelle, M.~Ranzato, R.~Hadsell, M.~Balcan, and H.~Lin,
  editors, {\em Advances in Neural Information Processing Systems}, volume~33,
  pages 7671--7684. Curran Associates, Inc., 2020.

\bibitem{alphacode}
Y.~Li, D.~Choi, J.~Chung, et~al.
\newblock Competition-level code generation with alphacode.
\newblock {\em Science}, 378(6624):1092--1097, 2022.

\bibitem{mcat_performance}
M.~Liu, T.~Okuhara, X.~Chang, R.~Shirabe, Y.~Nishiie, H.~Okada, and T.~Kiuchi.
\newblock Performance of chatgpt across different versions in medical licensing
  examinations worldwide: systematic review and meta-analysis.
\newblock {\em Journal of medical Internet research}, 26:e60807, 2024.

\bibitem{llm_statistical_causal_reasoning}
X.~Liu, Z.~Wu, X.~Wu, P.~Lu, K.-W. Chang, and Y.~Feng.
\newblock Are {LLM}s capable of data-based statistical and causal reasoning?
  benchmarking advanced quantitative reasoning with data.
\newblock In L.-W. Ku, A.~Martins, and V.~Srikumar, editors, {\em Findings of
  the Association for Computational Linguistics: ACL 2024}, pages 9215--9235,
  Bangkok, Thailand, Aug. 2024. Association for Computational Linguistics.

\bibitem{nonmyopic}
C.~Ma, H.~Zhao, J.~Zhang, J.~He, and L.~Kong.
\newblock Non-myopic generation of language models for reasoning and planning.
\newblock In {\em The Thirteenth International Conference on Learning
  Representations}, 2025.

\bibitem{gpt5}
{OpenAI}.
\newblock Introducing gpt-5.
\newblock \url{https://openai.com/index/introducing-gpt-5/}, Aug. 2025.

\bibitem{textquests}
L.~Phan, M.~Mazeika, A.~Zou, and D.~Hendrycks.
\newblock Textquests: How good are llms at text-based video games?, 2025.

\bibitem{poe}
W.~T. Piriyakulkij, Y.~Liang, H.~Tang, A.~Weller, M.~Kryven, and K.~Ellis.
\newblock Poe-world: Compositional world modeling with products of programmatic
  experts.
\newblock {\em CoRR}, abs/2505.10819, 2025.

\bibitem{reasoning_uncertainty}
M.~Pournemat, K.~Rezaei, G.~Sriramanan, A.~Zarei, J.~Fu, Y.~Wang,
  H.~Eghbalzadeh, and S.~Feizi.
\newblock Reasoning under uncertainty: Exploring probabilistic reasoning
  capabilities of llms, 2025.

\bibitem{raffin2021stablebaselines3}
A.~Raffin, A.~Hill, A.~Gleave, A.~Kanervisto, M.~Ernestus, and N.~Dormann.
\newblock Stable-baselines3: Reliable reinforcement learning implementations.
\newblock {\em Journal of Machine Learning Research}, 22(268):1--8, 2021.

\bibitem{random_shooting}
R.~Reiter, J.~Hoffmann, D.~Reinhardt, F.~Messerer, K.~Baumg{\"{a}}rtner,
  S.~Sawant, J.~Boedecker, M.~Diehl, and S.~Gros.
\newblock Synthesis of model predictive control and reinforcement learning:
  Survey and classification.
\newblock {\em CoRR}, abs/2502.02133, 2025.

\bibitem{ppo}
J.~Schulman, F.~Wolski, P.~Dhariwal, A.~Radford, and O.~Klimov.
\newblock Proximal policy optimization algorithms.
\newblock {\em arXiv preprint arXiv:1707.06347}, 2017.

\bibitem{shridhar2021alfworld}
M.~Shridhar, X.~Yuan, M.-A. Cote, Y.~Bisk, A.~Trischler, and M.~Hausknecht.
\newblock {\{}ALFW{\}}orld: Aligning text and embodied environments for
  interactive learning.
\newblock In {\em International Conference on Learning Representations}, 2021.

\bibitem{alphago}
D.~Silver, J.~Schrittwieser, K.~Simonyan, et~al.
\newblock Mastering the game of go without human knowledge.
\newblock {\em Nature}, 550(7676):354--359, 2017.

\bibitem{worldcoder}
H.~Tang, D.~Y. Key, and K.~Ellis.
\newblock Worldcoder, a model-based {LLM} agent: Building world models by
  writing code and interacting with the environment.
\newblock In {\em The Thirty-eighth Annual Conference on Neural Information
  Processing Systems}, 2024.

\bibitem{how_to_grow_a_mind}
J.~B. Tenenbaum, C.~Kemp, T.~L. Griffiths, and N.~D. Goodman.
\newblock How to grow a mind: Statistics, structure, and abstraction.
\newblock {\em Science}, 331(6022):1279--1285, 2011.

\bibitem{limit_planning}
K.~Valmeekam, M.~Marquez, S.~Sreedharan, and S.~Kambhampati.
\newblock On the planning abilities of large language models - a critical
  investigation.
\newblock In {\em Thirty-seventh Conference on Neural Information Processing
  Systems}, 2023.

\bibitem{alchemy}
J.~X. Wang, M.~King, N.~P.~M. Porcel, Z.~Kurth-Nelson, T.~Zhu, C.~Deck,
  P.~Choy, M.~Cassin, M.~Reynolds, H.~F. Song, G.~Buttimore, D.~P. Reichert,
  N.~C. Rabinowitz, L.~Matthey, D.~Hassabis, A.~Lerchner, and M.~Botvinick.
\newblock Alchemy: A benchmark and analysis toolkit for meta-reinforcement
  learning agents.
\newblock In {\em Thirty-fifth Conference on Neural Information Processing
  Systems Datasets and Benchmarks Track (Round 2)}, 2021.

\bibitem{scienceworld}
R.~Wang, P.~Jansen, M.-A. C{\^o}t{\'e}, and P.~Ammanabrolu.
\newblock {S}cience{W}orld: Is your agent smarter than a 5th grader?
\newblock In Y.~Goldberg, Z.~Kozareva, and Y.~Zhang, editors, {\em Proceedings
  of the 2022 Conference on Empirical Methods in Natural Language Processing},
  pages 11279--11298, Abu Dhabi, United Arab Emirates, Dec. 2022. Association
  for Computational Linguistics.

\bibitem{warrier2025benchmarkingworldmodellearning}
A.~Warrier, D.~Nguyen, M.~Naim, M.~Jain, Y.~Liang, K.~Schroeder, C.~Yang, J.~B.
  Tenenbaum, S.~Vollmer, K.~Ellis, and Z.~Tavares.
\newblock Benchmarking world-model learning, 2025.

\bibitem{grok4}
{xAI}.
\newblock Grok 4 model card.
\newblock Technical report, xAI, Aug. 2025.

\bibitem{revealing_planning}
J.~Xie, K.~Zhang, J.~Chen, S.~Yuan, K.~Zhang, Y.~Zhang, L.~Li, and Y.~Xiao.
\newblock Revealing the barriers of language agents in planning.
\newblock In L.~Chiruzzo, A.~Ritter, and L.~Wang, editors, {\em Proceedings of
  the 2025 Conference of the Nations of the Americas Chapter of the Association
  for Computational Linguistics: Human Language Technologies (Volume 1: Long
  Papers)}, pages 1872--1888, Albuquerque, New Mexico, Apr. 2025. Association
  for Computational Linguistics.

\bibitem{osworld}
T.~Xie, D.~Zhang, J.~Chen, X.~Li, S.~Zhao, R.~Cao, T.~J. Hua, Z.~Cheng,
  D.~Shin, F.~Lei, Y.~Liu, Y.~Xu, S.~Zhou, S.~Savarese, C.~Xiong, V.~Zhong, and
  T.~Yu.
\newblock {OSW}orld: Benchmarking multimodal agents for open-ended tasks in
  real computer environments.
\newblock In {\em The Thirty-eight Conference on Neural Information Processing
  Systems Datasets and Benchmarks Track}, 2024.

\bibitem{IVRE}
M.~Xu, G.~Jiang, W.~Liang, C.~Zhang, and Y.~Zhu.
\newblock Interactive visual reasoning under uncertainty.
\newblock In A.~Oh, T.~Naumann, A.~Globerson, K.~Saenko, M.~Hardt, and
  S.~Levine, editors, {\em Advances in Neural Information Processing Systems},
  volume~36, pages 42409--42432. Curran Associates, Inc., 2023.

\bibitem{qwen3}
A.~Yang, A.~Li, B.~Yang, B.~Zhang, B.~Hui, B.~Zheng, B.~Yu, C.~Gao, C.~Huang,
  C.~Lv, C.~Zheng, D.~Liu, F.~Zhou, F.~Huang, F.~Hu, H.~Ge, H.~Wei, H.~Lin,
  J.~Tang, J.~Yang, J.~Tu, J.~Zhang, J.~Yang, J.~Yang, J.~Zhou, J.~Zhou,
  J.~Lin, K.~Dang, K.~Bao, K.~Yang, L.~Yu, L.~Deng, M.~Li, M.~Xue, M.~Li,
  P.~Zhang, P.~Wang, Q.~Zhu, R.~Men, R.~Gao, S.~Liu, S.~Luo, T.~Li, T.~Tang,
  W.~Yin, X.~Ren, X.~Wang, X.~Zhang, X.~Ren, Y.~Fan, Y.~Su, Y.~Zhang, Y.~Zhang,
  Y.~Wan, Y.~Liu, Z.~Wang, Z.~Cui, Z.~Zhang, Z.~Zhou, and Z.~Qiu.
\newblock Qwen3 technical report, 2025.

\bibitem{webshop}
S.~Yao, H.~Chen, J.~Yang, and K.~R. Narasimhan.
\newblock Webshop: Towards scalable real-world web interaction with grounded
  language agents.
\newblock In A.~H. Oh, A.~Agarwal, D.~Belgrave, and K.~Cho, editors, {\em
  Advances in Neural Information Processing Systems}, 2022.

\bibitem{tree_of_thoughts}
S.~Yao, D.~Yu, J.~Zhao, I.~Shafran, T.~L. Griffiths, Y.~Cao, and K.~R.
  Narasimhan.
\newblock Tree of thoughts: Deliberate problem solving with large language
  models.
\newblock In {\em Thirty-seventh Conference on Neural Information Processing
  Systems}, 2023.

\bibitem{react}
S.~Yao, J.~Zhao, D.~Yu, N.~Du, I.~Shafran, K.~R. Narasimhan, and Y.~Cao.
\newblock React: Synergizing reasoning and acting in language models.
\newblock In {\em The Eleventh International Conference on Learning
  Representations}, 2023.

\bibitem{ying2025assessingadaptiveworldmodels}
L.~Ying, K.~M. Collins, P.~Sharma, C.~Colas, K.~I. Zhao, A.~Weller, Z.~Tavares,
  P.~Isola, S.~J. Gershman, J.~D. Andreas, T.~L. Griffiths, F.~Chollet, K.~R.
  Allen, and J.~B. Tenenbaum.
\newblock Assessing adaptive world models in machines with novel games, 2025.

\bibitem{DBLP:conf/nips/ZhaoLH23}
Z.~Zhao, W.~S. Lee, and D.~Hsu.
\newblock Large language models as commonsense knowledge for large-scale task
  planning.
\newblock In A.~Oh, T.~Naumann, A.~Globerson, K.~Saenko, M.~Hardt, and
  S.~Levine, editors, {\em Advances in Neural Information Processing Systems
  36: Annual Conference on Neural Information Processing Systems 2023, NeurIPS
  2023, New Orleans, LA, USA, December 10 - 16, 2023}, 2023.

\bibitem{webarena}
S.~Zhou, F.~F. Xu, H.~Zhu, X.~Zhou, R.~Lo, A.~Sridhar, X.~Cheng, Y.~Bisk,
  D.~Fried, U.~Alon, et~al.
\newblock Webarena: A realistic web environment for building autonomous agents.
\newblock {\em arXiv preprint arXiv:2307.13854}, 2023.

\bibitem{walle}
S.~Zhou, T.~Zhou, Y.~Yang, G.~Long, D.~Ye, J.~Jiang, and C.~Zhang.
\newblock Wall-e: World alignment by rule learning improves world model-based
  llm agents.
\newblock {\em arXiv preprint arXiv:2410.07484}, 2024.

\bibitem{walle2}
S.~Zhou, T.~Zhou, Y.~Yang, G.~Long, D.~Ye, J.~Jiang, and C.~Zhang.
\newblock {WALL-E} 2.0: World alignment by neurosymbolic learning improves
  world model-based {LLM} agents.
\newblock {\em CoRR}, abs/2504.15785, 2025.

\end{thebibliography}

\appendix

\section{Extended Related Works}\label{app:related_works}

\subsection{Benchmarks without Open-Ended Objectives}

In addition to those mentioned in the main text, another set of works is crafting and survival-based environments such as MineDojo \citep{minedojo}, HeroBench \citep{herobench}. All evaluate long-horizon, dependency-driven reasoning and require substantial game-specific knowledge to be successful. However, MineDojo and HeroBench are both deterministic and satisfaction based tasks. Only MineDojo involves spatial reasoning, and neither maps to real world tasks.

Instruction-following benchmarks such as BabyAI \citep{babyai} and AlfWorld \citep{shridhar2021alfworld} test navigation and object manipulation via natural language. BabyAI is grid-based, whereas AlfWorld simulates a household with common objects. These environments model tasks closer to the real world and are inherently multi-modal, yet still remain satisfaction-oriented in deterministic settings. 

\subsection{Benchmarks with Open-Ended Objectives}

Adding to the works discussed in the main paper, the Atari Learning Environment (ALE) \citep{ALE} has long served as a canonical human benchmark in AI. While impactful historically, it is now saturated, deterministic, and far removed from real-world domains.

Lastly, a unique benchmark is TextQuests \citep{textquests}. TextQuests probes LLMs in a zero-shot manner through interactive text-based adventures, scoring the model based on its progress through the game. It features open-ended action spaces and long horizons, but does not require learning on the fly, and the tasks are fully text-based and deterministic.

\section{Limitations}\label{sec:limitations}

Human players were evaluated in the unlimited resource setting, with full access to the documentation and unrestricted sandbox practice. Consequently, their scores serve as an aspirational upper bound rather than a directly comparable reference for the constrained evaluation conditions in which most AI agents are tested. 

Regarding agent memory, the ReAct baseline truncates context to the five most recent state--action pairs, which prevents agents from retaining information across the full episode. For the active world-model learning experiments (CH3), this is mitigated by instructing the agent to write notes for its future self; however, this approach has diminishing returns. Leaderboard data suggests that human players typically reach peak performance after five to six complete playthroughs---a learning horizon that exceeds what we evaluate. We ran preliminary experiments varying the days spend in sandbox learning and found that gains plateaued and then reversed, which we attribute to context bloat: as notes accumulate, redundant or contradictory entries increasingly crowd out useful information. Developing agents that can maintain compact, curated world models over extended interaction histories remains an open challenge.

Finally, our benchmark currently covers only easy and medium difficulty modes, and all evaluations use a fixed set of three test layouts. Although the layouts were withheld from agents during training, performance on a small number of maps may not fully characterise generalisation. Similarly, the RL baselines we evaluate are intentionally simple; more sophisticated RL approaches involving hierarchical or model-based methods may close the gap with LLM-based agents, and we leave this comparison to future work.

\section{Implementation Details}
\label{app:implementation_details}
\subsection{Evaluation Protocol}

Unless otherwise noted, all methods have full access to the game's documentation, which can be found in \cref{app:documentation}. Note that the section of the documentation on sandbox actions is omitted when the agent is not in sandbox mode.

In addition to conditioning on the documentation, we also perform experiments where the agent is allowed limited access to a sandbox learning mode prior to evaluation. In sandbox mode, the agents are given a budget of 100 standard \envname\ actions, and access to a number of sandbox actions. These sandbox actions include maximizing money, changing train layout, unlocking research etc... The full list of actions is outlined in the sandbox documentation reproduced in \cref{app:documentation}. To prevent infinite loops we impose a limit of 250 sandbox actions -- past this point sandbox actions are also deducted from the action budget. None of the agents tested hit this limit.

We evaluated all of our agents on 3 trajectories per testing layout. We reported the mean and standard deviation of the final park value (cash on hand, plus the value of physical assets and intellectual property -- see the game documentation in \cref{app:documentation} for details).

\subsection{AI System Evaluation}\label{app:ai_system}

To build a world model for \envacro\, we adapt Walle 1.0 \citep{walle}.  We modify all the original prompts to incorporate our new state and action specifications. Both the state and action specifications are represented using JSON objects. We further modify the state prediction prompt to incorporate and provide and examples of state transitions. For few shot examples, we associate transitions from the training data with an embedding of the action string, then at inference time we retrieve the top $k=5$ transitions based on the current action. We used the transitions seen in the sandbox learning experiments as our set of transitions; this ensures that the rule generation phase will be exposed to examples of failed actions that are specific to the model. For rule generation we follow \citep{walle}. We adopt two of the simplifications introduced in \cite{walle2}, namely going directly from generated natural language rules to generating code (i.e., eliminating the intermediate revision phase), and removing the ability of rules to invoke an LLM. The symbolic scenegraph introduced by \cite{walle2} is effectively provided by our state representation. 

For game playing experiments, and after conferring with the authors of WallE to best align with their approach, we combined the WallE world model with a ReACT \citep{react} agent using a Model Predicitive Control mechanism (MPC), specifically random shooting \citep{random_shooting} . I.e., actions are selected by selecting the best of $k$ rollouts guided by a world model. We select $k=5$ independent rollouts and roll out for 4 steps, re-sampling the action if the world-model predicts that the action is invalid (e.g., an invalid value for price, or attempting to place a building at coordinates that are outside of the bounds of the park). The ReAct history of the best action is is then used to initialize the history of the next $k$ rollouts. We use GPT-5-Nano as the underlying LLM policy for out MPC experiments.

\subsection{VLM ReAct details}
For the VLM experiments, we add the visual representation of the current park as an image to the message, keeping the remaining prompt and input observation identical. Compare to the fully GUI representation depicted in Appendix \cref{app:gui}, the provided image only contains the grid component of the park, as the remaining data panels are provided in more readable format through the JSON observation. To avoid unnecessary tokens, the message history does not contain the images, and only the current state is provided with an image.

\subsection{Reinforcement Learning Training}
\label{app:rl_training}

\subsection{Computational Resources}\label{app:cost}

All experiments can be performed on a standard consumer-grade laptop in combination with a cloud LLM inference provider. The RL experiments were performed on a single Amazon \texttt{t4g.xlarge} instance, with the longest experiments running over a period of 24 hours and costing 4 USD to run. The machine has 16GB memory and 4 vCPU cores, via AWS Graviton2 processors. To allow for parallelization of experiments, 4 Amazon \texttt{c6g.8xlarge} instances were used, each with 64GB memory and 32 vCPU cores, via AWS Graviton2 processors. LLM inference was via several providers through OpenRouter. The total cost of the experiments reported in the paper is approximately \$1500 USD; additional experiments were performed that did not make it into the final results; we estimate the cost of these experiments to be approximately \$500 USD. Individual experiment costs are reported in the relevant tables.

\paragraph{Algorithm.}
All reinforcement learning baselines are trained using Proximal Policy Optimization (PPO) \citep{ppo}, implemented in Stable-Baselines3 \citep{raffin2021stablebaselines3}. PPO is an on-policy actor--critic method that optimizes a clipped surrogate objective to ensure stable policy updates.

\paragraph{Observations.}
We support two observation modes. In \emph{full mode}, the observations contains the exact same information that the ReAct agents receive, but formatted in a method that enables consumption by more traditional machine learning algorithms. It consists of a dictionary combining a spatial grid representation ($20 \times 20 \times C$) with structured vector summaries encoding counts and attributes of rides, shops, staff, guests, and global park state (e.g., time, money, revenue). In \emph{simple mode}, the spatial grid is omitted and only the vector summaries are provided. 

\paragraph{Action Space and Hierarchical Policy.}
The action space is a high-dimensional \texttt{MultiDiscrete} space with hierarchical structure. Actions are decomposed into (i) an action type (e.g., place, move, remove, modify, research, survey, wait), followed by (ii) a sequence of parameters whose validity depends on previous choices (e.g., entity type, subtype, color class, spatial coordinates, or price). In \emph{full mode}, the agent must output all action parameters, matching the expressivity of the ReAct agents. In \emph{simple mode}, the (x, y) coordinates, price, and order quantity are filled in using simple heuristics, reducing the expressivity and complexity of the action space. The (x, y) coordinates use the same spatial heuristic developed for our spatial reasoning experiments. Price is always filled using max price. Order quantity is set to 2x the number of guests in the previous observation.

To handle this structure, we implement a custom hierarchical policy that samples decisions sequentially while applying dynamic action masking at each stage. Masks enforce feasibility constraints derived from the current state, including affordability, entity existence, and spatial validity (e.g., attractions must be placed on empty tiles adjacent to paths). This cascading masking enforces that actions are always valid, which significantly reduces the effective branching factor while preserving the original action semantics. Masks are computed using batched tensor operations to enable efficient GPU execution.

\paragraph{Optimization and Hyperparameters.}
Hyperparameters are selected using Bayesian optimization with Optuna \citep{akiba2019optuna}, using a Tree-structured Parzen Estimator (TPE) sampler \citep{bergstra2011algorithms}. The search space includes the learning rate, rollout length, number of optimization epochs, entropy regularization coefficient, GAE parameter $\lambda$, target KL threshold, value-function clipping, and number of minibatches. Underperforming trials are pruned early using a median-based pruner. We tune the hyperparameters for each combination of full/simple mode and difficulty.

We use a discount factor $\gamma = 0.995$, clipping range $0.2$, value-function loss coefficient $0.5$, and maximum gradient norm $0.5$.

\paragraph{Training and Early Stopping.}
Agents are trained for up to $100$ million environment steps using all five training layouts. Agents are periodically evaluated every $500{,}000$ environment steps on held-out layouts using mean episode reward. Training is terminated early if no improvement is observed for five consecutive evaluations after an initial burn-in period of $10$ million steps. The best-performing checkpoint is retained for final evaluation.

Training is performed using vectorized environments with multiple parallel workers. 

\paragraph{Rewards.}
Agents receive the environment’s intrinsic reward of differential park value. Rewards are normalized using a running mean and variance during training, which is necessary due to the large range that rewards can take on. Normalization is disabled during evaluation.

\section{Learning from the Documentation \& the Sandbox.}
\label{app:learning}

\begin{table}[h]
	\centering
	\caption{Results comparing models that were and were not provided the sandbox mode to learn in as well as the models that were and were not provided with the game's documentation. Score (\%) indicates the score relative to the human upper bound (unlimited resource setting), with 100.00\% indicating parity. SB: includes learning in sandbox mode. ND: model is not provided with the game's documentation.}
	\label{tab:learning}
	\begin{tabular}{llllll}
		\toprule
		Agent & LLM & Score (\%) & Runtime (s) & Cost (USD) & Train Cost \\
		\midrule
		\multicolumn{6}{c}{\textbf{EASY DIFFICULTY}} \\
		ReAct & GPT-5 Nano & 0.29$_{\pm 0.29}$ & 1,311$_{\pm 305}$ & 0.13$_{\pm0.02}$ & \$0.00 \\
		ReAct$_{\text{SB}}$ & GPT-5 Nano & 0.10$_{\pm 0.09}$ & 1,355$_{\pm 336}$ & 0.14$_{\pm0.02}$ & \$0.31 \\
		ReAct$_{\text{ND}}$ & GPT-5 Nano & 0.12$_{\pm 0.10}$ & 1,231$_{\pm 238}$ & 0.12$_{\pm0.01}$ & \$0.00 \\
		ReAct$_{\text{SB,ND}}$ & GPT-5 Nano & 0.09$_{\pm 0.09}$ & 1,369$_{\pm 358}$ & 0.13$_{\pm0.04}$ & \$0.35 \\
		ReAct & GPT-5 & 8.80$_{\pm 8.49}$ & 3,774$_{\pm 505}$ & 3.75$_{\pm0.32}$ & \$0.00 \\
		ReAct$_{\text{SB}}$ & GPT-5 & 8.70$_{\pm 7.53}$ & 3,655$_{\pm 346}$ & 4.16$_{\pm0.22}$ & \$10.26 \\
		ReAct$_{\text{ND}}$ & GPT-5 & 2.90$_{\pm 4.01}$ & 3,803$_{\pm 288}$ & 3.44$_{\pm0.28}$ & \$0.00 \\
		ReAct$_{\text{SB,ND}}$ & GPT-5 & 5.95$_{\pm 8.35}$ & 3,225$_{\pm 245}$ & 3.87$_{\pm0.33}$ & \$7.86 \\
		ReAct & Grok 4 & 2.82$_{\pm 4.50}$ & 2,908$_{\pm 437}$ & 6.60$_{\pm0.63}$ & \$0.00 \\
		ReAct$_{\text{SB}}$ & Grok 4 & 1.37$_{\pm 0.89}$ & 2,187$_{\pm 199}$ & 7.38$_{\pm0.51}$ & \$18.23 \\
		ReAct & Sonnet 4.5 & 2.28$_{\pm 2.02}$ & 937$_{\pm 124}$ & 6.06$_{\pm0.60}$ & \$0.00 \\
		ReAct$_{\text{SB}}$ & Sonnet 4.5 & 0.44$_{\pm 0.42}$ & 879$_{\pm 51}$ & 7.31$_{\pm0.55}$ & \$16.06 \\
		ReAct & Gemini 2.5 Pro & 1.82$_{\pm 2.78}$ & 787$_{\pm 84}$ & 3.06$_{\pm0.30}$ & \$0.00 \\
		ReAct$_{\text{SB}}$ & Gemini 2.5 Pro & 0.63$_{\pm 0.58}$ & 697$_{\pm 97}$ & 3.21$_{\pm0.29}$ & \$6.22 \\
		\multicolumn{6}{c}{\textbf{MEDIUM DIFFICULTY}} \\
		ReAct & GPT-5 Nano & 0.44$_{\pm 0.47}$ & 2,506$_{\pm 705}$ & 0.27$_{\pm0.04}$ & \$0.00 \\
		ReAct$_{\text{SB}}$ & GPT-5 Nano & 0.09$_{\pm 0.08}$ & 2,568$_{\pm 483}$ & 0.29$_{\pm0.03}$ & \$0.35 \\
		ReAct$_{\text{ND}}$ & GPT-5 Nano & 0.08$_{\pm 0.07}$ & 2,345$_{\pm 803}$ & 0.22$_{\pm0.08}$ & \$0.00 \\
		ReAct$_{\text{SB,ND}}$ & GPT-5 Nano & 0.17$_{\pm 0.23}$ & 2,532$_{\pm 440}$ & 0.27$_{\pm0.02}$ & \$0.30 \\
		ReAct & GPT-5 & 4.44$_{\pm 4.01}$ & 7,583$_{\pm 764}$ & 8.36$_{\pm0.48}$ & \$0.00 \\
		ReAct$_{\text{SB}}$ & GPT-5 & 6.52$_{\pm 4.81}$ & 7,897$_{\pm 822}$ & 9.16$_{\pm0.58}$ & \$9.82 \\
		ReAct$_{\text{ND}}$ & GPT-5 & 3.52$_{\pm 3.61}$ & 7,797$_{\pm 674}$ & 7.67$_{\pm0.75}$ & \$0.00 \\
		ReAct$_{\text{SB,ND}}$ & GPT-5 & 0.77$_{\pm 1.18}$ & 5,960$_{\pm 716}$ & 7.15$_{\pm0.73}$ & \$7.90 \\
		ReAct & Grok 4 & 0.44$_{\pm 0.16}$ & 5,438$_{\pm 683}$ & 13.00$_{\pm0.86}$ & \$0.00 \\
		ReAct$_{\text{SB}}$ & Grok 4 & 0.85$_{\pm 0.57}$ & 5,234$_{\pm 526}$ & 15.08$_{\pm1.05}$ & \$16.92 \\
		ReAct & Sonnet 4.5 & 1.06$_{\pm 0.80}$ & 1,713$_{\pm 86}$ & 12.19$_{\pm1.03}$ & \$0.00 \\
		ReAct$_{\text{SB}}$ & Sonnet 4.5 & 0.70$_{\pm 0.48}$ & 1,808$_{\pm 221}$ & 14.68$_{\pm0.90}$ & \$16.82 \\
		ReAct & Gemini 2.5 Pro & 0.52$_{\pm 0.22}$ & 1,423$_{\pm 126}$ & 5.97$_{\pm0.45}$ & \$0.00 \\
		ReAct$_{\text{SB}}$ & Gemini 2.5 Pro & 0.42$_{\pm 0.45}$ & 1,208$_{\pm 124}$ & 6.53$_{\pm0.46}$ & \$7.36 \\
		\bottomrule
	\end{tabular}
\end{table}

\cref{tab:learning} shows the results of the models under the standard set up (which includes the documentation, but not the sandbox learning) as well as models without the documentation (ND), and/or with the sandbox learning experience (SB). While there is a strong trend that reasoning models are able to effectively leverage the documentation to improve gameplay, the results are more mixed on the benefits of sandbox learning. Only GPT-5 (ND+easy, medium), GPT-5-Nano (ND+medium), and Grok-4 (medium) are ever able to derive an average benefit from sandbox learning, however this seems sensitive to game difficulty, with GPT-5 seemingly most reliably able to learn from the sandbox. While strong results cannot be drawn due to the high variance, this trend aligns with our qualitative inspection of the learnings generated. From human experience, this sandbox mode is incredibly useful, and we believe exploring how to effectively use these interactions is an exciting avenue for future work.

\FloatBarrier

\section{Future Work}\label{app:future_work}
We are actively developing \envacro\ to improve its use as an ongoing testbed for world modelling and human-AI benchmarking. To this end, we are developing a hard difficulty which will boast a horizon of 250 steps and feature at least three additional mechanics including: terraforming, guest preferences, and debt. Lastly, we will be expanding the observation options to include both a grid-vector based representation suitable for more traditional reinforcement learning, as well as an image-text based observation that is suitable for VLMS. Using these observations, we will benchmark an even larger set of existing AI systems.

\section{Detailed Model Evaluations}\label{app:detailedresults}

The absolute scores by human and AI systems are outlined in Tables \ref{tab:theislands} through \ref{tab:zig_zag}, for each of the 3 test layouts respectively.

\begin{table}[H]
	\centering
	\small
	\caption{Full experiment results with absolute scores for the test layout \texttt{the\_islands}. Average and standard deviation over 3 trials.}
	\label{tab:theislands}
	\begin{tabular}{llllll}
		\toprule
		Agent & LLM & Score & Runtime (s) & Cost (USD) & Train Cost \\
		\midrule
		\multicolumn{6}{c}{\textbf{EASY DIFFICULTY}} \\
		ReAct & GPT-5 Nano & 2,297$_{\pm 1,887}$ & 1,276$_{\pm 425}$ & 0.14$_{\pm0.03}$ & \$0.00 \\
		ReAct-MPC$_{\text{WALL-E}}$ & GPT-5 Nano & 397$_{\pm 89}$ & 27,939$_{\pm 2,186}$ & 7.54$_{\pm0.56}$ & \$2.47 \\
		ReAct-MPC$_{\text{Oracle}}$ & GPT-5 Nano & 18,142$_{\pm 11,333}$ & 9,755$_{\pm 877}$ & 3.77$_{\pm0.20}$ & \$0.00 \\
		ReAct$_{\text{SB}}$ & GPT-5 Nano & 1,751$_{\pm 2,454}$ & 1,329$_{\pm 485}$ & 0.15$_{\pm0.03}$ & \$0.31 \\
		ReAct$_{\text{ND}}$ & GPT-5 Nano & 1,361$_{\pm 1,815}$ & 1,105$_{\pm 241}$ & 0.12$_{\pm0.01}$ & \$0.00 \\
		ReAct$_{\text{SB,ND}}$ & GPT-5 Nano & 619$_{\pm 541}$ & 1,083$_{\pm 536}$ & 0.11$_{\pm0.07}$ & \$0.35 \\
		ReAct$_{\text{SpatialHeur}}$ & GPT-5 Nano & 5,469$_{\pm 7,752}$ & 1,352$_{\pm 297}$ & 0.15$_{\pm0.02}$ & \$0.00 \\
		ReAct & GPT-5 & 154,456$_{\pm 82,848}$ & 3,634$_{\pm 113}$ & 4.12$_{\pm0.13}$ & \$0.00 \\
		ReAct$_{\text{SB}}$ & GPT-5 & 209,816$_{\pm 82,785}$ & 3,838$_{\pm 531}$ & 4.42$_{\pm0.13}$ & \$10.26 \\
		ReAct$_{\text{ND}}$ & GPT-5 & 18,646$_{\pm 9,141}$ & 4,051$_{\pm 370}$ & 3.78$_{\pm0.21}$ & \$0.00 \\
		ReAct$_{\text{SB,ND}}$ & GPT-5 & 44,729$_{\pm 22,838}$ & 3,484$_{\pm 186}$ & 4.29$_{\pm0.07}$ & \$7.86 \\
		ReAct$_{\text{VLM}}$ & GPT-5 & 84,037$_{\pm 55,456}$ & 4,656$_{\pm 1,127}$ & 4.33$_{\pm0.13}$ & \$0.00 \\
		ReAct$_{\text{SpatialHeur}}$ & GPT-5 & 181,436$_{\pm 62,948}$ & 5,330$_{\pm 437}$ & 4.21$_{\pm0.15}$ & \$0.00 \\
		ReAct & Grok 4 & 122,871$_{\pm 190,128}$ & 3,111$_{\pm 352}$ & 7.39$_{\pm0.27}$ & \$0.00 \\
		ReAct$_{\text{SB}}$ & Grok 4 & 41,353$_{\pm 28,268}$ & 2,110$_{\pm 231}$ & 8.02$_{\pm0.25}$ & \$18.23 \\
		ReAct & Sonnet 4.5 & 72,763$_{\pm 61,477}$ & 946$_{\pm 141}$ & 6.77$_{\pm0.41}$ & \$0.00 \\
		ReAct$_{\text{SB}}$ & Sonnet 4.5 & 12,582$_{\pm 6,387}$ & 886$_{\pm 24}$ & 8.03$_{\pm0.10}$ & \$16.06 \\
		ReAct$_{\text{VLM}}$ & Sonnet 4.5 & 44,945$_{\pm 25,564}$ & 652$_{\pm 28}$ & 7.51$_{\pm0.09}$ & \$0.00 \\
		ReAct$_{\text{SpatialHeur}}$ & Sonnet 4.5 & 101,229$_{\pm 83,995}$ & 1,048$_{\pm 207}$ & 6.92$_{\pm0.43}$ & \$0.00 \\
		ReAct & Gemini 2.5 Pro & 70,543$_{\pm 107,185}$ & 782$_{\pm 89}$ & 3.35$_{\pm0.23}$ & \$0.00 \\
		ReAct$_{\text{SB}}$ & Gemini 2.5 Pro & 18,567$_{\pm 15,783}$ & 751$_{\pm 13}$ & 3.53$_{\pm0.06}$ & \$6.22 \\
		PPO$_{\text{Full}}$ & --- & 297$_{\pm 0}$ & 0$_{\pm 0}$ & 0.00$_{\pm0.00}$ & \$0.00 \\
		PPO$_{\text{Simple}}$ & --- & 87,029$_{\pm 6,925}$ & 0$_{\pm 0}$ & 0.00$_{\pm0.00}$ & \$0.00 \\
		ReAct & Qwen3 VL & 2,188$_{\pm 1,869}$ & 445$_{\pm 26}$ & 0.44$_{\pm0.02}$ & \$0.00 \\
		ReAct$_{\text{VLM}}$ & Qwen3 VL & 685$_{\pm 247}$ & 851$_{\pm 42}$ & 0.40$_{\pm0.01}$ & \$0.00 \\
		ReAct$_{\text{SpatialHeur}}$ & Qwen3 VL & 13,060$_{\pm 10,778}$ & 473$_{\pm 67}$ & 0.48$_{\pm0.04}$ & \$0.00 \\
		\midrule
		Human (Unlimited) &  & 2,455,681 &  &  &  \\
		\midrule
		\multicolumn{6}{c}{\textbf{MEDIUM DIFFICULTY}} \\
		ReAct & GPT-5 Nano & 4,367$_{\pm 6,992}$ & 2,040$_{\pm 854}$ & 0.26$_{\pm0.06}$ & \$0.00 \\
		ReAct-MPC$_{\text{Oracle}}$ & GPT-5 Nano & 93,305$_{\pm 37,018}$ & 24,186$_{\pm 5,739}$ & 9.20$_{\pm1.45}$ & \$0.00 \\
		ReAct$_{\text{SB}}$ & GPT-5 Nano & 733$_{\pm 801}$ & 2,184$_{\pm 626}$ & 0.29$_{\pm0.04}$ & \$0.35 \\
		ReAct$_{\text{ND}}$ & GPT-5 Nano & 3,895$_{\pm 3,406}$ & 1,938$_{\pm 1,401}$ & 0.20$_{\pm0.15}$ & \$0.00 \\
		ReAct$_{\text{SB,ND}}$ & GPT-5 Nano & 3,576$_{\pm 3,499}$ & 2,445$_{\pm 510}$ & 0.29$_{\pm0.02}$ & \$0.30 \\
		ReAct & GPT-5 & 79,194$_{\pm 25,890}$ & 7,117$_{\pm 278}$ & 8.91$_{\pm0.14}$ & \$0.00 \\
		ReAct$_{\text{SB}}$ & GPT-5 & 173,657$_{\pm 123,554}$ & 8,385$_{\pm 554}$ & 9.82$_{\pm0.32}$ & \$9.82 \\
		ReAct$_{\text{ND}}$ & GPT-5 & 57,091$_{\pm 23,314}$ & 7,834$_{\pm 472}$ & 8.12$_{\pm0.24}$ & \$0.00 \\
		ReAct$_{\text{SB,ND}}$ & GPT-5 & 11,114$_{\pm 6,942}$ & 6,442$_{\pm 610}$ & 7.92$_{\pm0.36}$ & \$7.90 \\
		ReAct$_{\text{VLM}}$ & GPT-5 & 137,227$_{\pm 66,693}$ & 10,064$_{\pm 1,053}$ & 10.08$_{\pm0.19}$ & \$0.00 \\
		ReAct & Grok 4 & 20,511$_{\pm 5,383}$ & 5,024$_{\pm 701}$ & 14.01$_{\pm0.36}$ & \$0.00 \\
		ReAct$_{\text{SB}}$ & Grok 4 & 15,225$_{\pm 5,083}$ & 5,533$_{\pm 283}$ & 16.44$_{\pm0.22}$ & \$16.92 \\
		ReAct & Sonnet 4.5 & 31,265$_{\pm 6,050}$ & 1,759$_{\pm 75}$ & 13.27$_{\pm0.36}$ & \$0.00 \\
		ReAct$_{\text{SB}}$ & Sonnet 4.5 & 25,384$_{\pm 24,292}$ & 1,710$_{\pm 89}$ & 15.73$_{\pm0.05}$ & \$16.82 \\
		ReAct$_{\text{VLM}}$ & Sonnet 4.5 & 37,660$_{\pm 21,139}$ & 1,311$_{\pm 54}$ & 15.26$_{\pm0.56}$  & \$0.00 \\
		ReAct & Gemini 2.5 Pro & 23,554$_{\pm 9,057}$ & 1,328$_{\pm 140}$ & 6.55$_{\pm0.12}$ & \$0.00 \\
		ReAct$_{\text{SB}}$ & Gemini 2.5 Pro & 8,598$_{\pm 2,834}$ & 1,258$_{\pm 119}$ & 6.98$_{\pm0.04}$ & \$7.36 \\
		PPO$_{\text{Full}}$ & --- & 18,965$_{\pm 12,868}$ & 0$_{\pm 0}$ & 0.00$_{\pm0.00}$ & \$0.00 \\
		PPO$_{\text{Simple}}$ & --- & 77,774$_{\pm 37,920}$ & 0$_{\pm 0}$ & 0.00$_{\pm0.00}$ & \$0.00 \\
		ReAct & Qwen3 VL & 7,828$_{\pm 2,391}$ & 1,674$_{\pm 119}$ & 0.85$_{\pm0.02}$ & \$0.00 \\
		ReAct$_{\text{VLM}}$ & Qwen3 VL & 11,786$_{\pm 1,380}$ & 1,381$_{\pm 272}$ & 0.91$_{\pm0.01}$ & \$0.00 \\
		ReAct$_{\text{SpatialHeur}}$ & Qwen3 VL & 23,625$_{\pm 33,615}$ & 735$_{\pm 90}$ & 1.00$_{\pm0.07}$ & \$0.00 \\
		\midrule
		Human (Unlimited) &  & 4,278,082 &  &  &  \\
		\bottomrule
	\end{tabular}
\end{table}

\clearpage

\newpage
\FloatBarrier
\begin{table}[H]
	\centering
	\small
	\caption{Full experiment results with absolute scores for the test layout \texttt{ribs}. Average and standard deviation over 3 trials.}
	\label{tab:ribs}
	\begin{tabular}{llllll}
		\toprule
		Agent & LLM & Score & Runtime (s) & Cost (USD) & Train Cost \\
		\midrule
		\multicolumn{6}{c}{\textbf{EASY DIFFICULTY}} \\
		ReAct & GPT-5 Nano & 2,068$_{\pm 1,194}$ & 1,329$_{\pm 388}$ & 0.13$_{\pm0.02}$ & \$0.00 \\
		ReAct-MPC$_{\text{WALL-E}}$ & GPT-5 Nano & 373$_{\pm 48}$ & 24,620$_{\pm 1,719}$ & 6.82$_{\pm0.18}$ & \$2.47 \\
		ReAct-MPC$_{\text{Oracle}}$ & GPT-5 Nano & 8,910$_{\pm 1,056}$ & 11,291$_{\pm 826}$ & 3.65$_{\pm0.14}$ & \$0.00 \\
		ReAct$_{\text{SB}}$ & GPT-5 Nano & 1,106$_{\pm 994}$ & 1,509$_{\pm 347}$ & 0.15$_{\pm0.02}$ & \$0.31 \\
		ReAct$_{\text{ND}}$ & GPT-5 Nano & 2,017$_{\pm 2,108}$ & 1,191$_{\pm 66}$ & 0.11$_{\pm0.01}$ & \$0.00 \\
		ReAct$_{\text{SB,ND}}$ & GPT-5 Nano & 1,208$_{\pm 469}$ & 1,615$_{\pm 27}$ & 0.15$_{\pm0.00}$ & \$0.35 \\
		ReAct$_{\text{SpatialHeur}}$ & GPT-5 Nano & 21,582$_{\pm 19,134}$ & 1,540$_{\pm 179}$ & 0.15$_{\pm0.01}$ & \$0.00 \\
		ReAct & GPT-5 & 168,727$_{\pm 117,429}$ & 4,281$_{\pm 617}$ & 3.63$_{\pm0.09}$ & \$0.00 \\
		ReAct$_{\text{SB}}$ & GPT-5 & 169,018$_{\pm 177,938}$ & 3,567$_{\pm 299}$ & 4.11$_{\pm0.03}$ & \$10.26 \\
		ReAct$_{\text{ND}}$ & GPT-5 & 91,328$_{\pm 81,325}$ & 3,726$_{\pm 178}$ & 3.32$_{\pm0.03}$ & \$0.00 \\
		ReAct$_{\text{SB,ND}}$ & GPT-5 & 190,074$_{\pm 167,726}$ & 3,158$_{\pm 142}$ & 3.73$_{\pm0.13}$ & \$7.86 \\
		ReAct$_{\text{VLM}}$ & GPT-5 & 155,304$_{\pm 247,464}$ & 3,551$_{\pm 238}$ & 3.80$_{\pm0.21}$ & \$0.00 \\
		ReAct$_{\text{SpatialHeur}}$ & GPT-5 & 445,397$_{\pm 105,250}$ & 5,303$_{\pm 903}$ & 3.82$_{\pm0.18}$ & \$0.00 \\
		ReAct & Grok 4 & 35,031$_{\pm 38,810}$ & 2,462$_{\pm 183}$ & 6.30$_{\pm0.11}$ & \$0.00 \\
		ReAct$_{\text{SB}}$ & Grok 4 & 12,786$_{\pm 6,894}$ & 2,328$_{\pm 144}$ & 7.20$_{\pm0.09}$ & \$18.23 \\
		ReAct & Sonnet 4.5 & 20,083$_{\pm 31,576}$ & 849$_{\pm 20}$ & 5.58$_{\pm0.13}$ & \$0.00 \\
		ReAct$_{\text{SB}}$ & Sonnet 4.5 & 2,304$_{\pm 802}$ & 833$_{\pm 49}$ & 6.87$_{\pm0.05}$ & \$16.06 \\
		ReAct$_{\text{VLM}}$ & Sonnet 4.5 & 1,801$_{\pm 591}$ & 613$_{\pm 22}$ & 6.05$_{\pm0.19}$ & \$0.00 \\
		ReAct$_{\text{SpatialHeur}}$ & Sonnet 4.5 & 121,672$_{\pm 67,600}$ & 924$_{\pm 53}$ & 5.99$_{\pm0.32}$ & \$0.00 \\
		ReAct & Gemini 2.5 Pro & 3,948$_{\pm 1,419}$ & 818$_{\pm 85}$ & 2.86$_{\pm0.11}$ & \$0.00 \\
		ReAct$_{\text{SB}}$ & Gemini 2.5 Pro & 6,823$_{\pm 6,700}$ & 714$_{\pm 56}$ & 3.12$_{\pm0.02}$ & \$6.22 \\
		PPO$_{\text{Full}}$ & --- & 15,221$_{\pm 3,160}$ & 0$_{\pm 0}$ & 0.00$_{\pm0.00}$ & \$0.00 \\
		PPO$_{\text{Simple}}$ & --- & 184,590$_{\pm 18,637}$ & 0$_{\pm 0}$ & 0.00$_{\pm0.00}$ & \$0.00 \\
		ReAct & Qwen3 VL & 5,818$_{\pm 4,866}$ & 471$_{\pm 45}$ & 0.36$_{\pm0.02}$ & \$0.00 \\
		ReAct$_{\text{VLM}}$ & Qwen3 VL & 2,464$_{\pm 2,832}$ & 975$_{\pm 144}$ & 0.34$_{\pm0.01}$ & \$0.00 \\
		ReAct$_{\text{SpatialHeur}}$ & Qwen3 VL & 7,565$_{\pm 5,616}$ & 298$_{\pm 12}$ & 0.38$_{\pm0.01}$ & \$0.00 \\
		\midrule
		Human (Unlimited) &  & 1,396,273 &  &  &  \\
		\midrule
		\multicolumn{6}{c}{\textbf{MEDIUM DIFFICULTY}} \\
		ReAct & GPT-5 Nano & 12,966$_{\pm 11,081}$ & 2,539$_{\pm 778}$ & 0.27$_{\pm0.06}$ & \$0.00 \\
		ReAct-MPC$_{\text{Oracle}}$ & GPT-5 Nano & 27,960$_{\pm 8,460}$ & 20,261$_{\pm 1,631}$ & 7.13$_{\pm0.45}$ & \$0.00 \\
		ReAct$_{\text{SB}}$ & GPT-5 Nano & 1,733$_{\pm 1,862}$ & 2,848$_{\pm 386}$ & 0.29$_{\pm0.03}$ & \$0.35 \\
		ReAct$_{\text{ND}}$ & GPT-5 Nano & 507$_{\pm 200}$ & 2,504$_{\pm 348}$ & 0.23$_{\pm0.02}$ & \$0.00 \\
		ReAct$_{\text{SB,ND}}$ & GPT-5 Nano & 992$_{\pm 205}$ & 2,495$_{\pm 510}$ & 0.26$_{\pm0.02}$ & \$0.30 \\
		ReAct & GPT-5 & 177,940$_{\pm 74,558}$ & 8,420$_{\pm 722}$ & 8.28$_{\pm0.22}$ & \$0.00 \\
		ReAct$_{\text{SB}}$ & GPT-5 & 240,658$_{\pm 27,591}$ & 6,901$_{\pm 50}$ & 8.87$_{\pm0.13}$ & \$9.82 \\
		ReAct$_{\text{ND}}$ & GPT-5 & 134,125$_{\pm 101,480}$ & 8,411$_{\pm 452}$ & 8.06$_{\pm0.65}$ & \$0.00 \\
		ReAct$_{\text{SB,ND}}$ & GPT-5 & 38,507$_{\pm 31,171}$ & 6,317$_{\pm 248}$ & 7.22$_{\pm0.22}$ & \$7.90 \\
		ReAct$_{\text{VLM}}$ & GPT-5 & 153,637$_{\pm 36,035}$ & 8,544$_{\pm 47}$ & 8.56$_{\pm0.16}$ & \$0.00 \\
		ReAct & Grok 4 & 11,523$_{\pm 1,717}$ & 5,252$_{\pm 425}$ & 12.80$_{\pm0.35}$ & \$0.00 \\
		ReAct$_{\text{SB}}$ & Grok 4 & 25,009$_{\pm 13,157}$ & 5,477$_{\pm 350}$ & 14.57$_{\pm0.26}$ & \$16.92 \\
		ReAct & Sonnet 4.5 & 24,141$_{\pm 23,098}$ & 1,660$_{\pm 74}$ & 11.56$_{\pm0.93}$ & \$0.00 \\
		ReAct$_{\text{SB}}$ & Sonnet 4.5 & 11,254$_{\pm 12,670}$ & 1,858$_{\pm 402}$ & 13.88$_{\pm0.71}$ & \$16.82 \\
		ReAct$_{\text{VLM}}$ & Sonnet 4.5 & 40,214$_{\pm 47,890}$ & 1,266$_{\pm 18}$ & 13.35$_{\pm1.42}$ & \$0.00 \\
		ReAct & Gemini 2.5 Pro & 7,995$_{\pm 4,373}$ & 1,454$_{\pm 111}$ & 5.58$_{\pm0.10}$ & \$0.00 \\
		ReAct$_{\text{SB}}$ & Gemini 2.5 Pro & 11,641$_{\pm 6,282}$ & 1,213$_{\pm 53}$ & 6.41$_{\pm0.03}$ & \$7.36 \\
		PPO$_{\text{Full}}$ & --- & 3,239$_{\pm 682}$ & 0$_{\pm 0}$ & 0.00$_{\pm0.00}$ & \$0.00 \\
		PPO$_{\text{Simple}}$ & --- & 81,985$_{\pm 6,626}$ & 0$_{\pm 0}$ & 0.00$_{\pm0.00}$ & \$0.00 \\
		ReAct & Qwen3 VL & 7,536$_{\pm 7,088}$ & 1,019$_{\pm 70}$ & 0.74$_{\pm0.01}$ & \$0.00 \\
		ReAct$_{\text{VLM}}$ & Qwen3 VL & 1,691$_{\pm 1,457}$ & 6,396$_{\pm 7,923}$ & 1.12$_{\pm0.64}$ & \$0.00 \\
		ReAct$_{\text{SpatialHeur}}$ & Qwen3 VL & 50,801$_{\pm 79,957}$ & 936$_{\pm 97}$ & 1.07$_{\pm0.24}$ & \$0.00 \\
		\midrule
		Human (Unlimited) &  & 1,979,867 &  &  &  \\
		\bottomrule
	\end{tabular}
\end{table}

\clearpage
\begin{table}[H]
	\small
	\centering
	\caption{Full experiment results with absolute scores for the test layout \texttt{zig\_zag}. Average and standard deviation over 3 trials.}
	\label{tab:zig_zag}
	\begin{tabular}{llllll}
		\toprule
		Agent & LLM & Score & Runtime (s) & Cost (USD) & Train Cost \\
		\midrule
		\multicolumn{6}{c}{\textbf{EASY DIFFICULTY}} \\
		ReAct & GPT-5 Nano & 2,960$_{\pm 1,381}$ & 1,328$_{\pm 193}$ & 0.13$_{\pm0.01}$ & \$0.00 \\
		ReAct-MPC$_{\text{WALL-E}}$ & GPT-5 Nano & 681$_{\pm 551}$ & 29,156$_{\pm 5,042}$ & 7.29$_{\pm0.85}$ & \$2.47 \\
		ReAct-MPC$_{\text{Oracle}}$ & GPT-5 Nano & 7,295$_{\pm 1,023}$ & 11,314$_{\pm 1,444}$ & 3.90$_{\pm0.51}$ & \$0.00 \\
		ReAct$_{\text{SB}}$ & GPT-5 Nano & 676$_{\pm 486}$ & 1,227$_{\pm 190}$ & 0.13$_{\pm0.01}$ & \$0.31 \\
		ReAct$_{\text{ND}}$ & GPT-5 Nano & 774$_{\pm 273}$ & 1,395$_{\pm 313}$ & 0.11$_{\pm0.01}$ & \$0.00 \\
		ReAct$_{\text{SB,ND}}$ & GPT-5 Nano & 829$_{\pm 570}$ & 1,410$_{\pm 92}$ & 0.13$_{\pm0.00}$ & \$0.35 \\
		ReAct$_{\text{SpatialHeur}}$ & GPT-5 Nano & 2,147$_{\pm 1,399}$ & 1,226$_{\pm 159}$ & 0.13$_{\pm0.01}$ & \$0.00 \\
		ReAct & GPT-5 & 38,655$_{\pm 64,599}$ & 3,407$_{\pm 107}$ & 3.50$_{\pm0.25}$ & \$0.00 \\
		ReAct$_{\text{SB}}$ & GPT-5 & 26,298$_{\pm 21,567}$ & 3,558$_{\pm 180}$ & 3.96$_{\pm0.13}$ & \$10.26 \\
		ReAct$_{\text{ND}}$ & GPT-5 & 6,672$_{\pm 303}$ & 3,632$_{\pm 137}$ & 3.23$_{\pm0.01}$ & \$0.00 \\
		ReAct$_{\text{SB,ND}}$ & GPT-5 & 11,667$_{\pm 6,144}$ & 3,032$_{\pm 151}$ & 3.58$_{\pm0.04}$ & \$7.86 \\
		ReAct$_{\text{VLM}}$ & GPT-5 & 12,525$_{\pm 7,495}$ & 3,996$_{\pm 175}$ & 3.76$_{\pm0.17}$ & \$0.00 \\
		ReAct$_{\text{SpatialHeur}}$ & GPT-5 & 26,673$_{\pm 14,533}$ & 4,970$_{\pm 325}$ & 3.62$_{\pm0.16}$ & \$0.00 \\
		ReAct & Grok 4 & 4,506$_{\pm 4,142}$ & 3,150$_{\pm 399}$ & 6.10$_{\pm0.20}$ & \$0.00 \\
		ReAct$_{\text{SB}}$ & Grok 4 & 7,325$_{\pm 5,060}$ & 2,123$_{\pm 199}$ & 6.92$_{\pm0.11}$ & \$18.23 \\
		ReAct & Sonnet 4.5 & 11,738$_{\pm 8,585}$ & 1,018$_{\pm 141}$ & 5.84$_{\pm0.27}$ & \$0.00 \\
		ReAct$_{\text{SB}}$ & Sonnet 4.5 & 3,089$_{\pm 3,191}$ & 917$_{\pm 45}$ & 7.02$_{\pm0.11}$ & \$16.06 \\
		ReAct$_{\text{VLM}}$ & Sonnet 4.5 & 9,114$_{\pm 13,706}$ & 652$_{\pm 29}$ & 6.23$_{\pm0.37}$ & \$0.00 \\
		ReAct$_{\text{SpatialHeur}}$ & Sonnet 4.5 & 10,832$_{\pm 9,517}$ & 934$_{\pm 61}$ & 5.87$_{\pm0.19}$ & \$0.00 \\
		ReAct & Gemini 2.5 Pro & 11,057$_{\pm 12,005}$ & 760$_{\pm 104}$ & 2.97$_{\pm0.29}$ & \$0.00 \\
		ReAct$_{\text{SB}}$ & Gemini 2.5 Pro & 3,159$_{\pm 3,920}$ & 625$_{\pm 147}$ & 2.99$_{\pm0.28}$ & \$6.22 \\
		PPO$_{\text{Full}}$ & --- & 1,961$_{\pm 151}$ & 0$_{\pm 0}$ & 0.00$_{\pm0.00}$ & \$0.00 \\
		PPO$_{\text{Simple}}$ & --- & 32,350$_{\pm 2,591}$ & 0$_{\pm 0}$ & 0.00$_{\pm0.00}$ & \$0.00 \\
		ReAct & Qwen3 VL & 2,784$_{\pm 586}$ & 670$_{\pm 95}$ & 0.36$_{\pm0.01}$ & \$0.00 \\
		ReAct$_{\text{VLM}}$ & Qwen3 VL & 1,131$_{\pm 760}$ & 629$_{\pm 23}$ & 0.36$_{\pm0.01}$ & \$0.00 \\
		ReAct$_{\text{SpatialHeur}}$ & Qwen3 VL & 2,409$_{\pm 1,078}$ & 354$_{\pm 24}$ & 0.39$_{\pm0.02}$ & \$0.00 \\
		\midrule
		Human (Unlimited) &  & 481,700 &  &  &  \\
		\midrule
		\multicolumn{6}{c}{\textbf{MEDIUM DIFFICULTY}} \\
		ReAct & GPT-5 Nano & 5,048$_{\pm 4,624}$ & 2,939$_{\pm 214}$ & 0.28$_{\pm0.02}$ & \$0.00 \\
		ReAct-MPC$_{\text{Oracle}}$ & GPT-5 Nano & 10,822$_{\pm 3,233}$ & 23,727$_{\pm 2,405}$ & 7.92$_{\pm1.14}$ & \$0.00 \\
		ReAct$_{\text{SB}}$ & GPT-5 Nano & 1,413$_{\pm 340}$ & 2,672$_{\pm 190}$ & 0.29$_{\pm0.02}$ & \$0.35 \\
		ReAct$_{\text{ND}}$ & GPT-5 Nano & 1,228$_{\pm 712}$ & 2,594$_{\pm 338}$ & 0.24$_{\pm0.03}$ & \$0.00 \\
		ReAct$_{\text{SB,ND}}$ & GPT-5 Nano & 3,345$_{\pm 3,011}$ & 2,658$_{\pm 467}$ & 0.26$_{\pm0.01}$ & \$0.30 \\
		ReAct & GPT-5 & 21,963$_{\pm 15,128}$ & 7,211$_{\pm 390}$ & 7.88$_{\pm0.22}$ & \$0.00 \\
		ReAct$_{\text{SB}}$ & GPT-5 & 29,821$_{\pm 28,585}$ & 8,406$_{\pm 395}$ & 8.77$_{\pm0.48}$ & \$9.82 \\
		ReAct$_{\text{ND}}$ & GPT-5 & 21,685$_{\pm 7,294}$ & 7,145$_{\pm 432}$ & 6.82$_{\pm0.37}$ & \$0.00 \\
		ReAct$_{\text{SB,ND}}$ & GPT-5 & 1,021$_{\pm 225}$ & 5,122$_{\pm 158}$ & 6.31$_{\pm0.06}$ & \$7.90 \\
		ReAct$_{\text{VLM}}$ & GPT-5 & 14,885$_{\pm 10,728}$ & 8,162$_{\pm 893}$ & 7.81$_{\pm1.00}$ & \$0.00 \\
		ReAct & Grok 4 & 2,379$_{\pm 590}$ & 6,038$_{\pm 587}$ & 12.17$_{\pm0.24}$ & \$0.00 \\
		ReAct$_{\text{SB}}$ & Grok 4 & 8,354$_{\pm 4,044}$ & 4,692$_{\pm 490}$ & 14.21$_{\pm0.19}$ & \$16.92 \\
		ReAct & Sonnet 4.5 & 10,868$_{\pm 8,456}$ & 1,720$_{\pm 106}$ & 11.73$_{\pm0.77}$ & \$0.00 \\
		ReAct$_{\text{SB}}$ & Sonnet 4.5 & 8,458$_{\pm 1,840}$ & 1,857$_{\pm 67}$ & 14.42$_{\pm0.12}$ & \$16.82 \\
		ReAct$_{\text{VLM}}$ & Sonnet 4.5 & 4,284$_{\pm 4,106}$ & 1,319$_{\pm 19}$ & 12.66$_{\pm0.12}$ & \$0.00 \\
		ReAct & Gemini 2.5 Pro & 5,344$_{\pm 2,266}$ & 1,486$_{\pm 104}$ & 5.80$_{\pm0.12}$ & \$0.00 \\
		ReAct$_{\text{SB}}$ & Gemini 2.5 Pro & 4,095$_{\pm 6,773}$ & 1,154$_{\pm 189}$ & 6.21$_{\pm0.60}$ & \$7.36 \\
		PPO$_{\text{Full}}$ & --- & 1,177$_{\pm 1,364}$ & 0$_{\pm 0}$ & 0.00$_{\pm0.00}$ & \$0.00 \\
		PPO$_{\text{Simple}}$ & --- & 2,434$_{\pm 3,502}$ & 0$_{\pm 0}$ & 0.00$_{\pm0.00}$ & \$0.00 \\
		ReAct & Qwen3 VL & 2,264$_{\pm 928}$ & 1,064$_{\pm 109}$ & 0.72$_{\pm0.04}$ & \$0.00 \\
		ReAct$_{\text{VLM}}$ & Qwen3 VL & 2,349$_{\pm 1,775}$ & 1,120$_{\pm 139}$ & 0.75$_{\pm0.01}$ & \$0.00 \\
		ReAct$_{\text{SpatialHeur}}$ & Qwen3 VL & 4,502$_{\pm 5,686}$ & 735$_{\pm 112}$ & 0.90$_{\pm0.20}$ & \$0.00 \\
		\midrule
		Human (Unlimited) &  & 887,772 &  &  &  \\
		\bottomrule
	\end{tabular}
\end{table}

\clearpage

\section{Qualitative Analysis of Learnings}
\label{app:learnings}
\begin{table}[h!]
	\centering
	\caption{A subset of learnings generated by GPT-5 during its 100 days in sandbox mode.}
	\label{tab:gpt5_learning}
	\begin{tabularx}{\linewidth}{p{0.04\linewidth} X p{0.31\linewidth}}
		\toprule
		\textbf{Step} & \textbf{Learning} & \textbf{Notes} \\
		\midrule
		4 & A single yellow ferris\_wheel priced near its max (\$4) attracted guests and generated revenue, and the ride and adjacent path accumulated dirt even without breakdowns—indicating that rides quickly impact cleanliness and likely benefit from early janitor staffing. & A simple, but useful observation \\
		10 & Hiring a blue mechanic restored ride uptime to near full and minimized out-of-service time, indicating mechanics quickly recover breakdowns and sustain revenue; rating impacts can lag a day due to start-of-day calculation. & Identifies lag in park rating bump, however everything here is already noted in the game documentation. \\
		11 & Park rating improved the day after uptime stabilized (all rides at 1.0) following hiring a mechanic, confirming rating effects can lag and that sustained uptime materially lifts rating and revenue. & Confirms the above, but again no new information garnered. \\
		19 & After adding a high-capacity ride, drink demand exceeded 300/day and the yellow drink shop still stocked out, while increasing the yellow food shop’s order\_quantity to 120 eliminated stockouts but left 23 units unsold—confirming that zero-cost drink shops should be overstocked aggressively, whereas food shops with item costs should be tuned close to observed demand to minimize waste. & Useful to note that the zero-cost drink shops should be overstocked, but is surrounded by detail that is much too specific that specific day. \\
		43 & After adding distributed drink coverage and letting a full day pass, drink visits per guest rose (1.36), but higher footfall strained cleanliness (ride/path cleanliness dipped to 0.82–0.98) and uptime on high-breakdown rides sagged—showing that as capacity and traffic grow, janitorial coverage must scale to maintain rating and guest happiness. & Again filled with overly specific details. Noting scaling can be useful, but could be inferred simply from the documentation. \\
		48 & Fresh survey results (with cleanliness and uptime stabilized) still show many exits due to ``Too few unique drink/food shops'' and ``Too few unique attractions'', indicating that distributing additional shops across high-traffic corridors and adding more attractions are key levers for raising happiness, rating, and value. & Positive use of guest surveys to get more information. But the conclusions it draws are weak: drink and food directly impact thirst and hunger, which, if too high, can impact happiness; impact on value is even more indirect. \\
		53 & Increasing the ATM's order quantity eliminated its stockouts and the park rating rose the next day, confirming specialty shops can bottleneck guest spending and that tuning their inventory supports rating and value. & Poor causal understanding. ATMs impact money spent, not park rating. It likely over emphasizes the impact of its action rather than the state of the park. Park ratings can pendulum: a dirty park will lower park rating leading to fewer guests leading to a cleaner park leading higher park rating again. Over generalizes to specialty shops, but each specialty shop is unique. \\
		77 & After a day, the new green food shop (price 18) maintained 1.0 uptime but sold only ~50–61 items against an order\_quantity of 200 (item cost 4), indicating significant daily waste; tuning nonzero-cost shop inventory to observed demand should lift profit without hurting uptime. & Noting that stocking accurately is useful, but it does not include any actionable learnings. \\
		88 & The newly placed red food shop at max price achieved solid same-day demand (~50 sales) with 1.0 uptime; because its items have nonzero cost, over-ordering (e.g., 80 with ~30 leftover) wastes money, so inventory should be tuned close to observed demand. & Repetitive learning from day 43 (not the only example of this) \\
		\bottomrule
		
	\end{tabularx}
\end{table}

Summarizing the analysis in \cref{tab:gpt5_learning}, we see that GPT-5 is able to identify some simple, but useful trends: janitors and mechanics are important early, park rating can lag a day, and stocking should be considered carefully. However, most of these can be directly inferred from the documentation with some thought. Further, the useful trends are surrounded by very specific information about that specific day and the learnings are also not particularly actionable beyond their most straightforward application. Additionally, when encountering more complicated relationships, it over-generalizes and does a poor job of untangling causal effects. Instead it presumes that its action is the main driver of all change. It successfully explores different components of the game, from yellow carousels to red food shops, but spends a lot of time testing high-level subclasses, without understanding that it will not have access to the full range of subclasses. Perhaps most important to note here, is that it does not try to provide any information about which higher subclasses are more valuable, which would be invaluable in guiding research during the evaluation stage. 

\begin{table}[h!]
	\centering
	\caption{A subset of learnings generated by Claude Sonnet 4.5 during its 100 days in sandbox mode.}
	\label{tab:claude_sonnet}
	\begin{tabularx}{\linewidth}{p{0.04\linewidth} X p{0.31\linewidth}}
		\toprule
		\textbf{Step} & \textbf{Learning} & \textbf{Notes} \\
		\midrule
		7 & Adding a food shop near rides improved rating 11.36 $\rightarrow$ 46.02; guests visited 0.35 food shops on average $\rightarrow$ meeting hunger needs boosts rating and satisfaction. & While the general trend that food shops can help is true, the rating gap is way too large to be attributed to a single food shop. It is much more likely to be due to the rubber effect of cleanliness. \\
		12 & A yellow mechanic (repair rate 2) took a full day (998 steps) to repair a red roller coaster (\$100 000) $\rightarrow$ low-tier mechanics too slow for expensive rides. & True, but an odd experiment to run, and this information is available in the documentation. \\
		20 & A green mechanic (repair rate 20) maintained red roller coaster at 61\% uptime; revenues \$4150, costs \$1600 + \$3780 $\rightarrow$ -\$165 profit $\rightarrow$ even high tiers struggle to make it profitable. & Overly specific to a specific context, but also omits the necessary context. Fails to understand that stochasticity can be the cause of the low uptime for this specific turn. \\
		29 & Park with 4 diverse rides, 2 shops, 2 staff $\rightarrow$ consistent \$1088 profit, 69 guests, rating 36.53 $\rightarrow$ ride diversity + inventory management + mechanic = stable profit. & Reasonable and useful assessment. \\
		32 & Adding too many same rides hurts returns $\rightarrow$ 3rd yellow carousel $\rightarrow$ guests 102$\rightarrow$87, profit 1681$\rightarrow$888, rating 37.49$\rightarrow$31.62 $\rightarrow$ diversity $>$ capacity for profitability. & Again the claim can be true, but the data here is misleading. Adding a new ride can only increase the park rating in that turn (on a longer horizon, the happiness of guests may be impacted leading to a lower park rating). Drop in park rating is more likely due to the cleanliness. \\
		46 & Green mechanic (repair rate 20) maintained blue coaster 91\%, fixed others to 99/97\%, 980 steps @ \$980 cost, profit 5214$\rightarrow$3993 $\rightarrow$ effective but expensive maintenance. & Extremely specific. \\
		91 & Once repaired, blue mechanic (repair rate 8) maintains 100\% uptime at \$100/day salary (no repair costs) $\rightarrow$ excellent value for mid-tier maintenance. & Potentially useful, the causally incorrect. 100\% can only occur because of the lack of a breakdown, which is a stochastic event.  \\
		\bottomrule
	\end{tabularx}
\end{table}

Analyzing \cref{tab:claude_sonnet}, we see that Claude Sonnet 4.5 tends to be extremely specific on changes in numbers, but does not properly consider factors outside of its most immediate action. It also does not consider stochasticity or other potential causal links in its insights, leading to several spurious correlations.

\begin{table}[h!]
	\centering
	\caption{A subset of learnings generated by Grok-4 during its 100 days in sandbox mode.}
	\label{tab:grok_learning}
	\begin{tabularx}{\linewidth}{p{0.04\linewidth} X p{0.31\linewidth}}
		\toprule
		\textbf{Step} & \textbf{Learning} & \textbf{Notes} \\
		\midrule
		1 & Water tiles adjacent to a ride increase its excitement by 1 per adjacent water tile, as evidenced by a red roller coaster with base excitement 10 achieving 12 when adjacent to two water tiles. Placing a single high-intensity ride attracts a limited number of guests up to its capacity but results in low park rating and some cleanliness decline without staff or diversity. & Useful information, but information that exists in the documentation. Interesting to start learning with the red roller coaster, which is very difficult to attain. \\
		4 & Without janitors, ride cleanliness can drop to critically low levels over multiple days, resulting in zero guest usage of the ride despite available capacity, as evidenced by cleanliness reaching 0.0 and no guests entertained in a day with high guest turnover due to “too few rides.” & Repeating information that can be found in the manual \\
		19 & A yellow specialist (clown) continues to boost guest happiness in lines over multiple days, as evidenced by entertaining 351 guests in the second day (up from 280), correlating with improved park rating from 15.91 to 30.25, increased guests from 79 to 142, and higher average rides visited from 0.93 to 1.35. & Useful insight that is not directly in the manual. \\
		20 & Attractions cannot be placed on tiles that already contain a path; attempting to do so results in an invalid action error. They must be on empty tiles adjacent to paths. & Again, this is in the manual. \\
		40 & Same constraint as Step 20—attempting to place attractions on path tiles results in invalid actions. & Repeating its own mistakes. \\
		68 & Hiring a blue specialist (stocker) at a valid path location successfully places it, but it may not perform restocking or movement if no shops drop below threshold during the day, as evidenced by success\_metric\_value 0.0 and tiles\_traversed 0 in the initial day. & Found in the documentation and generally uninformative. \\
		71 & Removing a ride of the same subtype as an existing one reduces excitement, capacity, guests, and rating, but can increase avg rides visited per guest and profit, e.g., excitement 28.16 $\rightarrow$ 22.56, capacity 88 $\rightarrow$ 76, guests 114 $\rightarrow$ 84, rating 24.18 $\rightarrow$ 19.18, avg rides visited 1.07 $\rightarrow$ 1.46, and profit -18 880 $\rightarrow$ +6 966 after selling a duplicate roller coaster. & Poor causal and long-horizon understanding. The large swing in profit likely arises from buying a ride one turn (leading to a high cost), and then selling it the next (high revenue). This short-sided analysis shows the opposite of the long term impact. \\
		75 & Guest surveys with one ride reveal ``Too few unique attraction'' and sometimes ``Too few drink shops,'' with thirst 0.4-0.7 and varying happiness; 22/25 results cite uniqueness issues. & Positive use of guest surveys, but does not leverage the information toward actionable learnings. \\
		80 & Hiring a yellow mechanic to repair a broken high-value ride (red roller coaster) performs $\sim$1000 repair steps at cost $\sim$1000 in one day but does not immediately restore uptime (0), showing that one mechanic is insufficient. & Over generalization. One yellow mechanic is insufficient for a red roller coaster. \\
		81 & A yellow mechanic continues repairs over multiple days ($\sim$1000 steps/day @ 1000 cost/day) but still 0 uptime after 2 days $\rightarrow$ shows slow repair rate. & Correct, but uniformative. \\
		82 & Adding a red mechanic assists repairs (2450 steps + yellow 98) $\rightarrow$ repair completed in 1 day, uptime 0.9, guests return; shows multi-mechanic synergy. & Useful, but three days were spent to learn what is largely intuitive given the documentation. \\
		\bottomrule
	\end{tabularx}
\end{table}

Condensing the analysis in \cref{tab:grok_learning}, we see that Grok-4 does a very poor job of learning useful actionable insights. It spends many turns to identify knowledge that readily available in the manual. Like GPT-5, it also struggles with identifying causal relations and over attributes outcomes to the most recent action at hand, without consideration to the existing park layout. Lastly, it appears to spend very little time learning about the early stages of the game, which are critical to reach the later states it does explore.

\begin{table}[h!]
	\centering
	\caption{A subset of learnings generated by Gemini 2.5-Pro during its 100 days in sandbox mode.}
	\label{tab:gemini_learning}
	\begin{tabularx}{\linewidth}{p{0.04\linewidth} X p{0.31\linewidth}}
		\toprule
		\textbf{Step} & \textbf{Learning} & \textbf{Notes} \\
		\midrule
		6 & Placing a food shop effectively addresses hunger, increases spend per guest, and generates profit. & Useful, but already in the documentation. \\
		7 & Placing a drink shop satisfies thirst and generates revenue, but an order\_quantity of 50 was insufficient, causing stockouts. High-excitement rides (e.g., blue roller coaster) break down often. & Useful, but already in the documentation. \\
		8 & A high-tier (red) mechanic can repair a broken ride within one day, drastically improving uptime and revenue potential. & Somewhat useful, but fails to note the high price of red mechanics and the research required to get there. \\
		9 & Increasing a shop’s order\_quantity via modify prevents stockouts, ensuring sustained revenue. Research on roller coasters completed, unlocking red subclass. & In the documentation, or unhelpful overly specific information. \\
		48 & With park growth, two red janitors may not maintain cleanliness and three red mechanics may struggle $\rightarrow$ min uptime 0.75, min cleanliness 0.62. & It misses crucial context: with 20 red rides, this may very well be true. With a few yellow rides, two red janitors is excessive. \\
		55 & Setting research speed = fast unlocks next tier (blue) in 1 day but at high cost; high traffic increases dirtiness (cleanliness 0.65 even with 3 janitors). & Useful to note the timing of research, but fast research is too expensive to be used frequently. Testing the speeds for slow or medium would likely be more useful. \\
		57 & Upgrading a high-demand shop (drink) to higher tier (blue) dramatically boosts rating, guests, and revenue, but guest surge lowers cleanliness. & Upgrade insight can be gleaned from the documentation. Noting guest surge can be useful, if it uses that to pre-emptively hire janitors (which it won't do). \\
		74 & Hiring 7 red mechanics + 6 red janitors $\rightarrow$ restores uptime 0.97 and strong profit $\rightarrow$ optimal staffing ratio for large park. & Again, it omits critical context here. \\
		94 & Hiring a 12th red mechanic did not solve uptime (0.93) and caused -12,749 loss $\rightarrow$ more mechanics $\neq$ more profitability. & Again, it omits critical context here.  \\
		\bottomrule
	\end{tabularx}
\end{table}

Finally, from the analysis in \cref{tab:gemini_learning}, we see that Gemini also does a very poor job of learning useful actionable insights. It largely regurgitates information that is already found in the documentation; on few instances when it tries to identify more specific trends, it omits necessary context.

\clearpage

\section{Qualitative Analysis of Spatial Placement}
\label{app:spatial}
\begin{figure}[h!]
	\centering
	\begin{center}
		\begin{subfigure}[t]{0.45\linewidth}
			\centering
			\includegraphics[trim={0 0 900pt 300pt},clip,width=\linewidth]{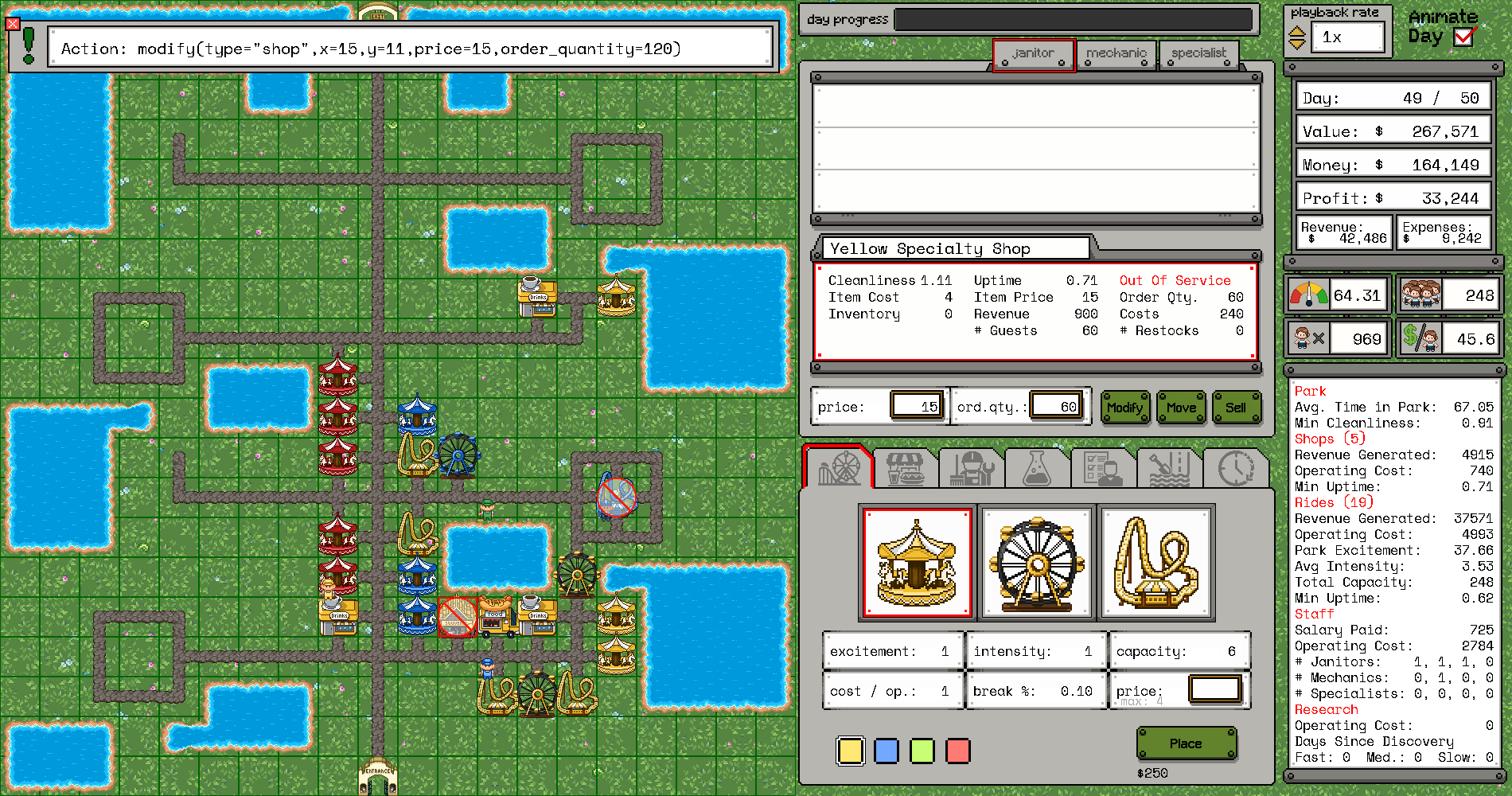}
			\caption{Final park of GPT-5's best playthrough on Ribs Easy.}
		\end{subfigure}
		\quad
		\begin{subfigure}[t]{0.45\linewidth}
			\centering
			\includegraphics[trim={0 0 900pt 300pt},clip,width=\linewidth]{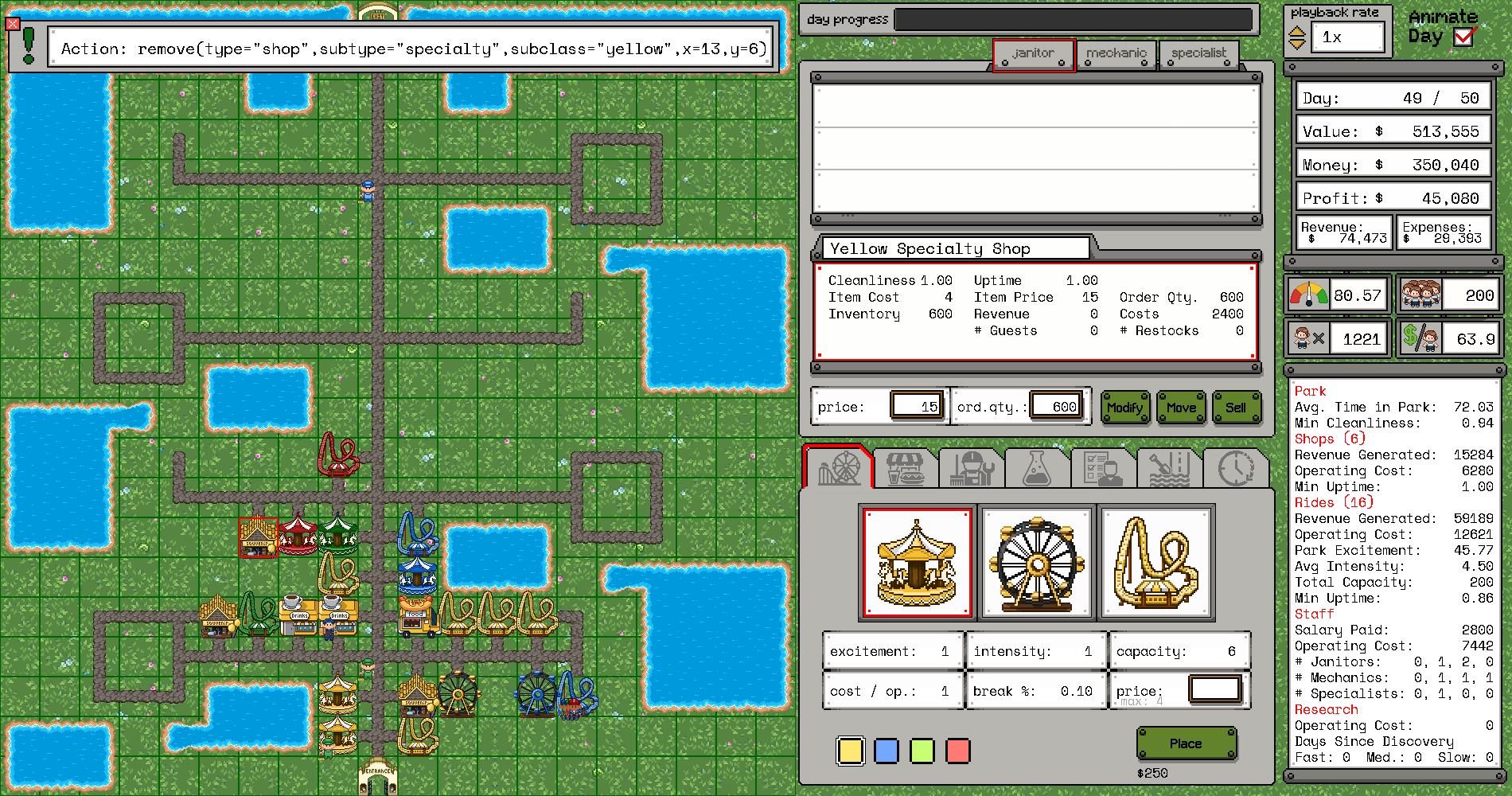}
			\caption{Final park of GPT-5 with the added spatial heuristic's best playthrough on Ribs Easy.}
		\end{subfigure}
	\end{center}
	
	\vspace{0.5em}
	
	\centering
	\begin{center}
		\begin{subfigure}[t]{0.45\linewidth}
			\centering
			\includegraphics[trim={0 0 900pt 300pt},clip,width=\linewidth]{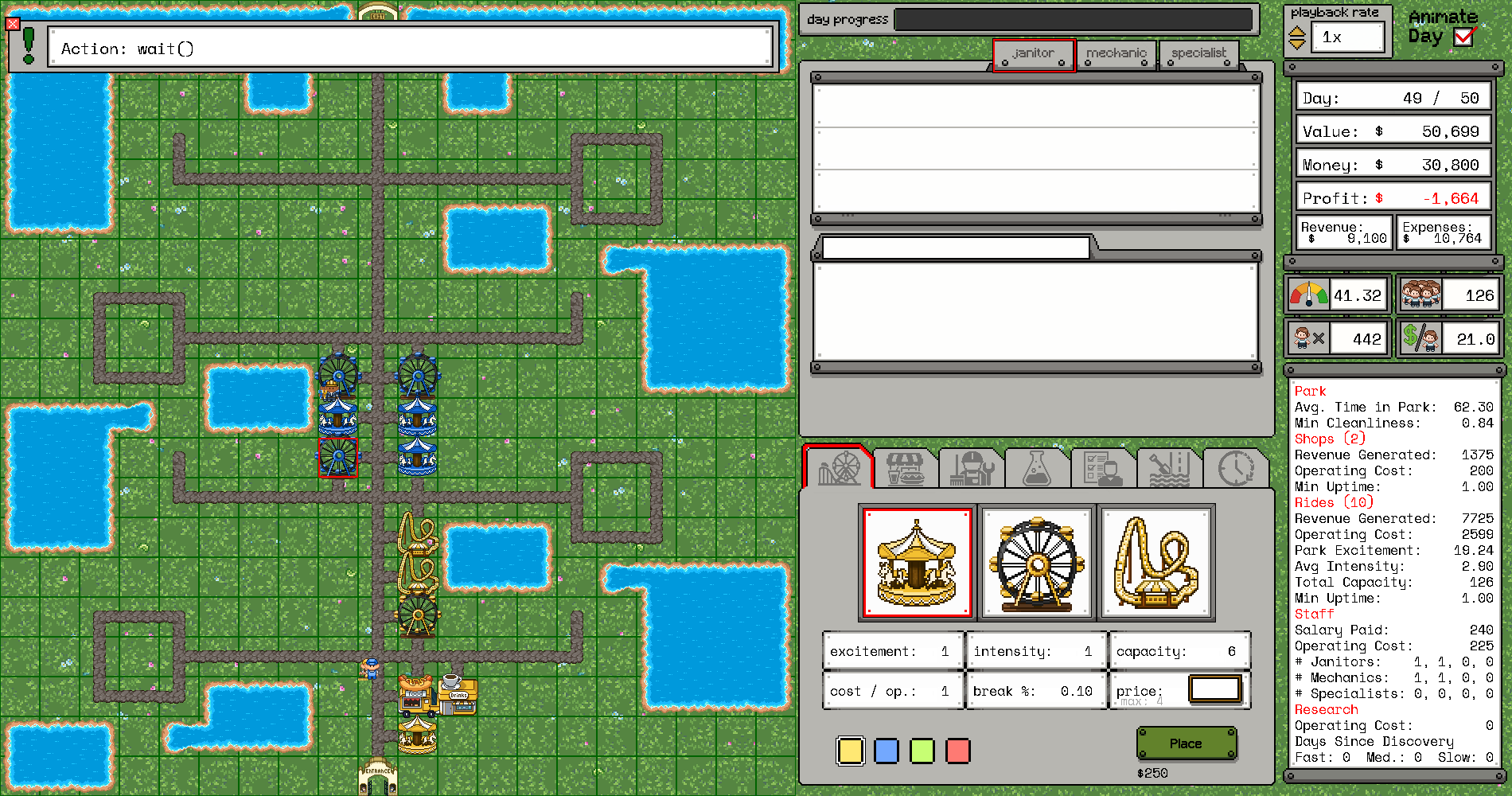}
			\caption{Final park of Claude Sonnet 4.5's best playthrough on Ribs Easy.}
		\end{subfigure}
		\quad
		\begin{subfigure}[t]{0.45\linewidth}
			\centering
			\includegraphics[trim={0 0 900pt 300pt},clip,width=\linewidth]{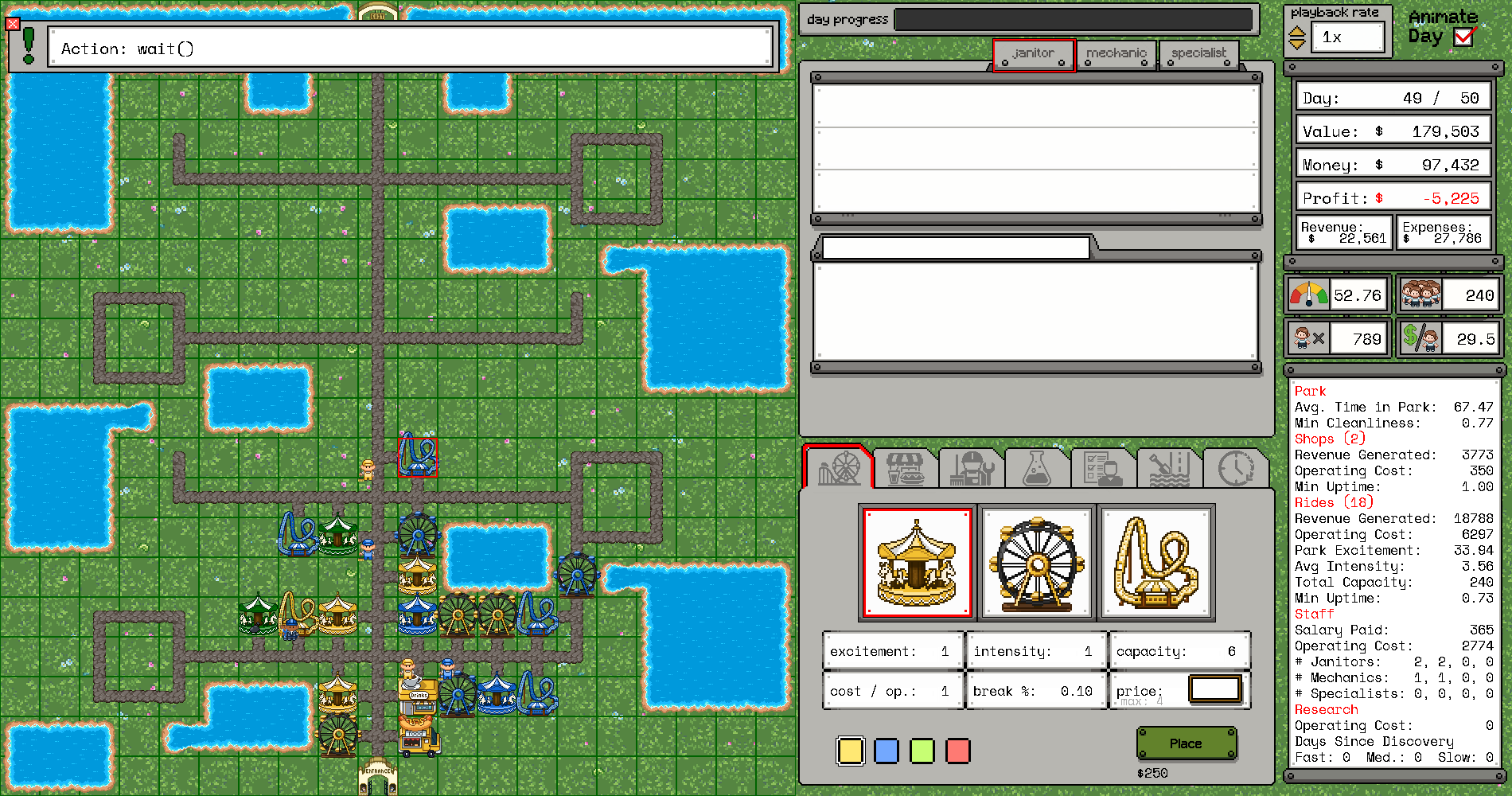}
			\caption{Final park of Claude Sonnet 4.5 with the added spatial heuristic's best playthrough on Ribs Easy.}
		\end{subfigure}
	\end{center}
	
	\caption{Comparison of the park design with and without the added spatial heuristic overriding the LLMs placement choices.}
	\label{fig:spatial heuristic qualitative analysis}
\end{figure}

\FloatBarrier

\section{Qualitative Analysis of Failures and Successes}\label{app:qual_succ}

Manually reviewing our trajectories, we were able to various trends, some general and only relevant to specific models. Starting with the latter, we identified that
Grok-4 is strongly biased towards surveying guests. In some cases this allows it to address guest concerns brought to its attention (e.g., by hiring a clown or constructing a drink shop). However, due to the relatively high cost of surveying a guest (500 per guest, twice the price of the cheapest ride), its tendency to survey frequently (e.g. once every 5 turns) rapidly drains its coffers.

A related failure mode that ensnares both Grok-4 and Gemini is prematurely launching research. This choice to invest in research rather than expanding the base park results in slower growth. Compounding this issue, Grok-4 is sometimes unable to track what it is researching, and so performs a single step of research, accumulates funds over several turns, and then performs a second step of research on a different topic. This prevents it from making progress.

Sonnet (and to a lesser extent Grok-4) tend to be overly reactive and risk averse with building placement. Both, not uncommonly, construct and then immediately sell off an attraction -- this results in a loss of 33\% of the value which could be prevented by either avoiding the initial construction, or by experimenting with building placement or ticket pricing. In one case, Grok-4 placed a souvenir shop near the entrance (an excellent placement as souvenir shops are high profit margin and rely on foot traffic -- which will be high near the entrance), but immediately sold it citing ``The specialty shop added at (17,10) is underutilized (only 0.06 average visits, 2 guests served) but incurred high stocking costs (\$172 operating, only \$30 revenue), leading to negative profit last day. Removing it will recover 66\% of build cost (\$165) and eliminate ongoing losses''. This was overly reactive, as the stock could have been reduced instead.

Closely related, Sonnet (and to a lesser extend Gemini) suffers a failure where instead of performing the free action of moving a ride, they instead sell and then construct an attraction.

Sonnet tends to play conservatively; in some cases prematurely concluding that expansion is not possible (due to its poor placement combined with a tendency to panic and act reactively), and instead decides to wait until the conclusion of the game. 

With sandbox learning, we observed that Sonnet struggled with the longer context. Specifically, partway through six of the nine runs it would repeatedly fail to format its output correctly, and ultimately decide that waiting until the game was completed was its best option. In one trajectory we saw Gemini similarly enter a loop where it repeatedly failed to format its output correctly, but it was able to recover after 4 failed attempts and continued play normally.

In \envacro\ the yellow specialist (i.e., clown) is an employee that improves guest happiness in lines -- only indirectly improving revenue by increasing park rating (leading to more guests in the future), and by making guests stay and spend in the park for longer. We observe that Sonnet never hires clowns, and Gemini only rarely. Grok-4 hires clowns extensively (as mentioned, sometimes in response to a guest survey highlighting guest happiness as an issue), though sometimes enters over-hiring (e.g., about 5 clowns in one of the runs -- it did correct this by firing some of the clowns). GPT-5 also typically hires clowns, further improving its performance compared to other models.

GPT-5 is a stronger player than the other models tested; its shortcomings are more subtle. While it's generally quite good at building near the entrance and clustered, it still fails when faced with the subtleties of paths and connectivity. For example, on the \texttt{zig\_zag} layout it will build attractions that are nearby as the crow flies, but quite far apart for guests due to the winding path connecting the attractions. Similarly, on \texttt{the\_islands} it consistently initially fails to understand that different islands are not connected by paths until after it has constructed attractions in inaccessible locations. It is generally able to recover and move these attractions. 

GPT-5 plays most aggressively of the models studied, building new attractions, in some cases as much as 10 duplicate rides (typically yellow or blue roller coasters). This aggression -- combined with its competent placement -- gives it stronger revenue in the early game making research more practical. However, GPT-5 tends to begin by researching blue roller coasters, the most expensive blue unlock. This causes it to lose time once the research is complete (e.g., in one evaluation there were 16 turns between the unlocking of the blue coaster and its first construction). In most cases, once it has performed its research it only builds blue coasters -- preferring continued aggressive expansion with blue coasters over ride diversification. This also leads to problems with park management: due to the high guests counts the yellow staff are unable to keep the park clean or the rides maintained. This results in bimodal park behaviour where guests attendance oscillates -- high attendance leads to dirtiness leading to low attendance the next day leading to a maintainable janitorial load leading to high attendance the next day. Instead of performing research into better staff to address this issue, GPT-5 continues aggressive expansion, ignoring the reduction in visitors (as well as the penalties for duplicated rides). Relating to this preference for immediate expansion over all else, in some evaluations GPT-5 never performs research, instead only building yellow attractions.  %
Interestingly, we observe that with sandbox learning GPT-5 is much more willing to perform research -- sometimes researching staff (the only agent studied to do so), and researching blue ferris wheels in 50\% of the evaluations.

\subsection{GPT-5 Sandbox Learning without Documentation}

As to why sandbox learning decreases performance in medium difficulty when the documentation is removed, the cause is unclear. The notes still explore various tiers of attractions, and appear similar to those learned under different conditions. Studying the medium trajectories with sandbox learning, GPT-5 tends to construct a ferris wheel very early in the game -- a poor choice as when guest counts are low the high-capacity ferris wheel will have long lines. Combined with the increased impulsiveness when documentation is removed, this results in GPT-5 repeatedly constructing and demolishing attractions for under performing. Why this occurs is unclear. Reviewing the learnings we found one encouraging the construction of yellow ferris wheels: \texttt{Adding even a low-tier unique ride (yellow ferris wheel) near a secondary hub increased park rating without hurting uptime or cleanliness, reinforcing that additional attractions (not just high-tier) can lift value when operations are stable}. This is largely true, though only if guest counts are sufficiently high. Given that this is the only relevant learning, it is unclear if this is responsible for the behaviour observed. If so, then this would suggest that online learning would benefit future \envacro\ agents.

\section{Observation Example}
\label{app:obs_example}
\begin{jsonblock}
	{
		"available_entities": {
			"carousel": ["yellow"], "drink": ["yellow"], "ferris_wheel": ["yellow"], "food": ["yellow"], "janitor": ["yellow"], "mechanic": ["yellow"], "roller_coaster": ["yellow"], "specialist": ["yellow"], "specialty": ["yellow"]},
		"entrance": [0,1],
		"exit": [19,18],
		"expenses": 122,
		"fast_days_since_last_new_entity": 0,
		"guest_survey_results": {"age_of_results": 5, "list_of_results": []},
		"guests": {
			"avg_drink_shops_visited": 0.59,
			"avg_food_shops_visited": 0.64,
			"avg_money_spent": 6.95,
			"avg_rides_visited": 0.5,
			"avg_specialty_shops_visited": 0.0,
			"avg_time_in_park": 113.36,
			"total_guests": 30
		},
		"horizon": 100,
		"medium_days_since_last_new_entity": 0,
		"min_cleanliness": 0.85,
		"money": 387,
		"new_entity_available": false,
		"parkId": "263dde2b-95b5-4a7a-b6a6-c7c15d50843b",
		"park_rating": 31.68,
		"paths": [
		{
			"cleanliness": 0.98,
			"x": 0,
			"y": 2
		},
		...
		{
			"cleanliness": 0.96,
			"x": 19,
			"y": 17
		}
		],
		"profit": 91,
		"research_operating_cost": 0,
		"research_speed": "none",
		"research_topics": ["carousel", "ferris_wheel", "roller_coaster", "drink", "food", "specialty", "janitor", "mechanic", "specialist"],
		"revenue": 213,
		"rides": {
			"avg_intensity": 1.0,
			"min_uptime": 1.0,
			"ride_list": [
			{
				"avg_guests_per_operation": 1.07,
				"avg_wait_time": 12.13,
				"breakdown_rate": 0.001,
				"capacity": 6,
				"cleanliness": 0.87,
				"cost_per_operation": 1,
				"excitement": 1,
				"guests_entertained": 16,
				"intensity": 1,
				"operating_cost": 15,
				"out_of_service": false,
				"revenue_generated": 64,
				"subclass": "yellow",    
				"subtype": "carousel",
				"ticket_price": 4,
				"times_operated": 15,
				"uptime": 1.0,
				"x": 0,
				"y": 4
			}
			],
			"total_capacity": 6,
			"total_excitement": 0.8,
			"total_operating_cost": 15,
			"total_revenue_generated": 64,
			"total_rides": 1
		},
		"shops": {
			"min_uptime": 1.0,
			"shop_list": [
			{
				"cleanliness": 0.92,
				"guests_served": 18,
				"inventory": 82,
				"item_cost": 0,
				"item_price": 3,
				"number_of_restocks": 0,
				"operating_cost": 0,
				"order_quantity": 100,
				"out_of_service": false,
				"revenue_generated": 54,
				"subclass": "yellow",
				"subtype": "drink",
				"uptime": 1.0,
				"x": 0,
				"y": 5
			},
			{
				"cleanliness": 0.94,
				"guests_served": 19,
				"inventory": 1,
				"item_cost": 1,
				"item_price": 5,
				"number_of_restocks": 0,
				"operating_cost": 20,
				"order_quantity": 20,
				"out_of_service": false,
				"revenue_generated": 95,
				"subclass": "yellow",
				"subtype": "food",
				"uptime": 1.0,
				"x": 1,
				"y": 2
			}
			],
			"total_operating_cost": 20,
			"total_revenue_generated": 149,
			"total_shops": 2
		},
		"slow_days_since_last_new_entity": 0,
		"staff": {
			"staff_list": [
			{
				"operating_cost": 2,
				"salary": 25,
				"subclass": "yellow",
				"subtype": "janitor",
				"success_metric": "amount_cleaned",
				"success_metric_value": 0.05,
				"tiles_traversed": 1,
				"x": 0,
				"y": 4
			},
			{
				"operating_cost": 0,
				"salary": 60,
				"subclass": "yellow",
				"subtype": "specialist",
				"success_metric": "guests_entertained",
				"success_metric_value": 97.0,
				"tiles_traversed": 2,
				"x": 0,
				"y": 4
			}
			],
			"total_janitors": [1, 0, 0, 0],
			"total_mechanics": [0, 0, 0, 0],
			"total_operating_cost": 2,
			"total_salary_paid": 85,
			"total_specialists": [1, 0, 0, 0],
		},
		"step": 5,
		"value": 750,
		"waters": [
		{
			"x": 2,
			"y": 4
		},
		{
			"x": 2,
			"y": 5
		},
		...
		{
			"x": 16,
			"y": 18
		}
		]
	}
	
\end{jsonblock}

\section{Prompt Templates}\label{app:prompts}

\subsection{Base Prompt Templates}\label{app:basepromptteplates}

For all models we begin with these base templates and evaluate on a test layout for 3 steps to check for any errors in model behaviour. These base ReAct prompts are modified from LangChain\footnote{\tiny{\url{https://python.langchain.com/api_reference/_modules/langchain/agents/react/agent.html}}}. We budget 30 minutes per model for prompt tuning to correct any observed misbehaviour on the part of the model. 

State observations are provided as json strings, with any error messages from the \envname\ environment or the world model appended to the state along with the last action taken; e.g., \texttt{NOTE: While attempting the action `place(x=12, y=9, type="shop", subtype="drink", subclass="yellow", price=3, order\_quantity=-1)` the error `\{'message': 'Inventory order\_quantity cannot be negative: -1', 'type': 'invalid\_action'\}` occurred.}

\begin{lstlisting}[caption={ReAct System Prompt Template}]
	You are playing a game where you are the manager of a theme park. 
	You can only take a single action each day. 
	Your objective is to maximize park value over a {horizon} day period.
	Note that you are playing in *{difficulty} difficulty*.
	
	Use the following format to structure your output:
	
	Thought: you should always think about what to do
	Action: the action to take, should be one of the following: {actions_list}
	Action Input: the input to the action, as correctly formatted list of python arguments -- e.g.: arg1="hi", arg2=34
	Observation: the user will provide you with result of the action
	... (this Thought/Action/Action Input/Observation repeats until the game is complete)
	
	Below is the documentation of the game:
	
	{GAMEPLAY_RULES}
\end{lstlisting}

\begin{lstlisting}[caption={ReAct User Prompt Template}]
	{REACT_HISTORY}
	
	Observation: {state_with_error_msg}
	Question: What is the best action I should take now to take to maximize park value?
	Thought:
\end{lstlisting}

In sandbox learning the following prompt was used:

\begin{lstlisting}[caption={Sandbox Learning ReAct System Prompt}]
	Instructions:
	You are tasked with playing an amusement park simulator.
	Before starting the layouts that you will be evaluated on, 
	You will have {max_sandbox_steps} in-game days in a sandbox mode to explore and interact with the environment.
	Use this period to learn how the game works and experiment with different actions and strategies.
	During this sandbox mode, you will have access to several additional commands:
	undo_day, max_money, max_research, reset, and switch_layouts.
	These are designed to help you learn the games dynamics more efficiently.
	These actions do not count toward your learning budget.
	However, they will not be available during the evaluation phase.
	You are free to learn on any of the provided layouts.
	
	After completing the sandbox stage, you will begin the evaluation.
	This will occur on a layout not available during sandbox mode.
	During evaluation, your goal will be to maximize the park's value over a {horizon} day period.
	
	You will be playing on {difficulty} difficulty.
	
	Documentation:
	A detailed guide for the game has been provided.
	Before playing, please read through the documentation carefully.
	The documentation contains useful information, rules, and tips to help you succeed.
	
	Strictly obey the following format to structure your output:
	
	Learnings Summary: a one or two sentence summary of the useful information you have learned from the previous interaction (if anything). This information should be general enough to apply even on a game layout with a different arrangement of paths. This information should be about normal gameplay -- i.e., not specific to sandbox mode. Remember, this is for only information *supported by evidence* (no unbased speculation and no reasoning about your next action!) that will be useful when you play again in the future. Your learnings from all turns will be made available to you, both during exploration and when you're evaluated.
	Thought: you should always think about what to do
	Action: the action to take, should be one of the normal game actions ({actions_list}) or the sandbox game actions ({sandbox_actions_list})
	Action Input: the input to the action, as correctly formatted list of python arguments -- e.g.: arg1="hi", arg2=34
	Observation: the user will provide you with result of the action
	... (this Thought/Action/Action Input/Observation repeats until the game is complete)
	
	
	For example:
	
	Learnings Summary: I can increase park value by...
	Thought: Based on this, I should try to...
	Action: place
	Action Input: x=5, y=12, type="ride", subtype="carousel", subclass="yellow", price=3
	
	
	Below is the documentation for the game:
	
	{SANDBOX_GAMEPLAY_RULES}
\end{lstlisting}

\begin{lstlisting}[caption={Sandbox Learning ReAct User Prompt}]
	# Collated Past Learnings Summaries
	{LEARNINGS_LIST_TO_DATE}
	
	# Recent Actions
	
	{REACT_HISTORY}
	
	# Current State
	
	Observation: {state_with_error_msg}
	Question: What is the best action you can take next to explore the environment with the aim of learning how to maximize park value? You have {sandbox_steps_left} non-sandbox actions remaining.
	Learnings Summary:
\end{lstlisting}

\begin{lstlisting}[caption={ReAct System Prompt Augmented with Learnings}]
	You are playing a game where you are the manager of a theme park. 
	You can only take a single action each day. 
	Your objective is to maximize park value over a {horizon} day period.
	Note that you are playing in *{difficulty} difficulty*.
	
	Use the following format to structure your output:
	
	Thought: you should always think about what to do
	Action: the action to take, should be one of the following: {actions_list}
	Action Input: the input to the action, as correctly formatted list of python arguments -- e.g.: arg1="hi", arg2=34
	Observation: the user will provide you with result of the action
	... (this Thought/Action/Action Input/Observation repeats until the game is complete)
	
	Below is the documentation of the game:
	
	{GAMEPLAY_RULES}
	
	You previously had the chance to explore the game and make notes. These are the notes you made:
	
	Step0: {STEP_0_LEARNING}
	
	Step1: {STEP_1_LEARNING}
	
	Step2: {STEP_2_LEARNING}
	
	...
	
	Step100: {STEP_100_LEARNING}
\end{lstlisting}

The prompt for sandbox GPT-5 Nano was modified to better separate learnings from thoughts about next steps.

The prompt for  Claude Sonnet 4.5 was modified to discourage the model from hallucinating next state observations.

\section{Human GUI}\label{app:gui}

\begin{figure*}[h!]
	\centering
	\includegraphics[width=0.75\linewidth]{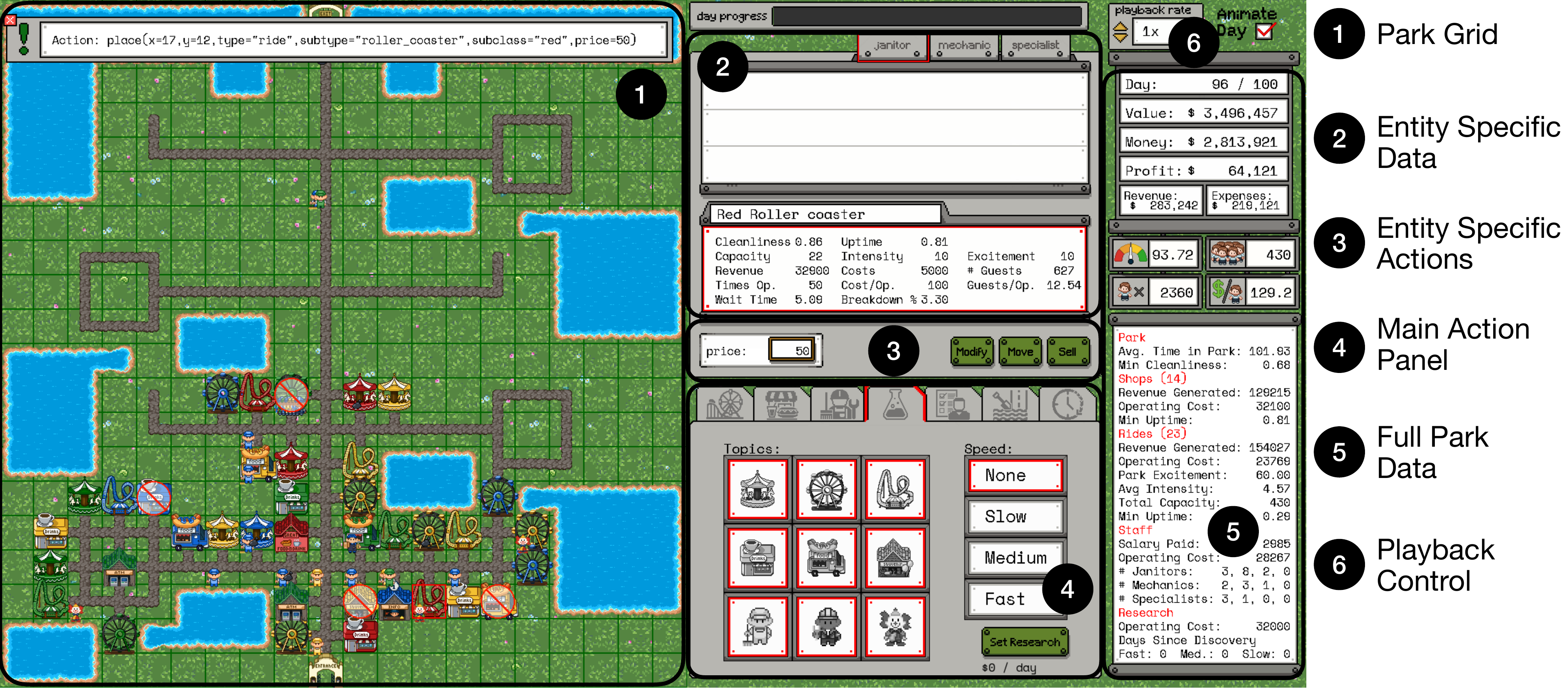}
	\caption{The GUI view of \envacro.}
	\label{fig:main_fig}
\end{figure*}

The GUI for human use is illustrated in Fig~\ref{fig:main_fig}. This is the GUI available on our web leaderboard. Note that we did not perform a human study, instead we provided a leaderboard that is open to the public. No crowdsourcing or contract work was used. No personally identifiable information is collected, and only final game score is reported. A review was performed, and no risks to participants were found. A screenshot of the interface is provided in Figure~\ref{fig:main_fig}, and the full game manual is available at Appendix~\ref{app:documentation}.

\subsection{EULA for Score Submission}\label{app:eula}

\begin{lstlisting}[caption={Leaderboard EULA for score submission}]
LEADERBOARD DATA CONSENT & PRIVACY NOTICE

By choosing to submit your score to the leaderboard, you agree that limited gameplay data (such as your score, duration, and in-game actions) and a leaderboard name of your choice will be collected and displayed publicly.

This data is used solely to operate and maintain the leaderboard, detect unfair play,  improve the game experience, and conduct aggregate analyses comparing human performance to artificial intelligence (AI) agents.

We do not collect personal information such as real names or email addresses. All analyses use anonymous or aggregated data only. Your leaderboard entry (leaderboard name and score) will be visible to other players while the leaderboard is active.

You may request removal of your leaderboard entry at any time by contacting us at [REDACTED FOR DOUBLE BLIND REVIEW]

Submitting your score is optional. By clicking the checkbox below and confirming submission, you consent to the collection and public display of your gameplay data and leaderboard name as described above.

Leaderboard Moderation

We reserve the right to remove any user or entry from the leaderboard at our discretion,  for any reason, including but not limited to inappropriate or offensive leaderboard names, suspected cheating, data manipulation, or other behavior deemed inconsistent with fair play and community standards.

Last updated: November 10, 2025

\end{lstlisting}

\section{Game Documentation}
\label{app:documentation}
The following pages include the game's full documentation.


\section*{Supplementary material}

# Game Mechanics

The goal of the game is to maximize your amusement park's value within a set number of days. As the manager of the park, you perform one action at the start of each day. The park then opens, and guests interact with the park for a full day.

There are three difficulty modes in the game (*Easy*, *Medium*, and *Hard*):

- **Easy:** All attractions are available from the beginning. Short time horizon (50 days).
- **Medium:** Only yellow attractions are available from the beginning; other attractions must be researched. Medium time horizon (100 days).
- **Hard:** Coming soon.

There are seven primary components to the game. We provide a high-level overview here, followed by detailed descriptions of each:

- **The Park:** Defined by a square grid (defaults to 20x20) and contains all other components. The amusement park has an entrance and an exit connected by a path. Your park also has a *rating* that reflects guest satisfaction, park upkeep, and overall park quality.
- **Terrain:** There are three kinds of terrain -- Paths, Water, and Empty.
- **Rides:** Rides are one of two types of attractions you can place in your amusement park. They are the core of the park and draw guests in. There are three subtypes of rides: Carousels, Ferris Wheels, and Roller Coasters.
- **Shops:** Shops are the second type of attraction. They allow you to cater to your guests' needs. There are three subtypes of shops: Drink, Food, and Specialty. Drink shops quench guests' thirst; Food shops satisfy their hunger; Specialty shops provide unique services.
- **Staff:** Staff can be hired to maintain your park. There are three subtypes of staff: Janitors, Mechanics, and Specialists. Janitors keep the park clean, Mechanics repair rides, and Specialists perform a variety of support tasks.
- **Subclasses & Research:** Each ride, shop, and staff subtype has four subclasses—yellow, blue, green, and red. In Easy mode these are unlocked from the start; in Medium and Hard modes, higher subclasses must be unlocked through research.
- **Guests:** The people your park is built for! Guests interact with your park, spending money to purchase ride tickets, food, drinks, and more.

---

## Terrain

- **Empty tiles:** Blank tiles on which something can be built.
- **Path tiles:** Used by guests to move around your park. All attractions must be placed on an empty tile adjacent to a path tile.
- **Water tiles:** Each water tile adjacent to a ride increases that ride's excitement by 1.

In Hard mode, terrain can be terraformed. Paths can be built ($1000) and removed ($2500). Water tiles can similarly be added ($5000) and removed ($10000).

---

## Rides

Rides are the core of your amusement park. They are the primary driver in determining how many guests visit your park, they improve guest happiness, and contribute to the overall value of the park. Rides have several key attributes:

- **Capacity:** How many guests can fit on the ride at one time. Capacity also affects how frequently the ride operates. The cumulative capacity of your park is also a key factor in how many guests visit the park.
- **Excitement:** How thrilling the ride is. Higher excitement scores increase guest happiness and park rating.
- **Intensity:** How intense the ride experience is. Keeping the average intensity balanced (around 5) ensures your park caters to a wide range of guests and improves your rating.
- **Ticket price:** The amount guests must pay to ride. If a guest cannot afford the ticket, they are rejected, which decreases their happiness.
- **Cost per operation:** The amount it costs each time the ride runs.

  **TIP:** Building multiple rides of the same kind (i.e., identical subtype and subclass) yields diminishing returns for your park rating. Diversifying your attractions will lead to a higher overall rating.

Rides only operate if guests have boarded. After the first guest boards, rides wait a few turns to allow more guests to join. This waiting time is longer for rides with higher capacity but decreases as more guests board. A full ride always operates immediately.

Rides may break down after operating. When broken, they are out of service and will turn away guests, negatively affecting their happiness and your park rating.

  **TIP:** Hire mechanics to fix broken rides promptly.

There are three subtypes of rides:

- **Carousels:** Cheap to build and operate, and they rarely break down. However, they provide limited excitement and capacity.
- **Ferris Wheels:** Intermediate rides with the highest capacities among all subtypes.
- **Roller Coasters:** Expensive but high-value rides that offer the highest excitement and intensity scores, at the cost of frequent breakdowns.

For the exact parameters of each ride, see [All Rides](#).

---

## Shops

Shops are necessary to adequately cater to guest needs, and provide additional value to a park. Shops have several key attributes:

- **Order quantity:** The amount of inventory purchased at the start of each day. Unsold inventory is lost at the end of the day.
- **Item cost:** The cost of purchasing one unit of inventory.
- **Item price:** The price guests pay for one unit of inventory.

At the start of each day, shops are stocked according to their order quantity at the item cost. If there are insufficient funds to fully restock all shops, partial stocking will occur. If a shop runs out of inventory during the day, it will go out of service, turning away guests and lowering their happiness and your park rating.

  **TIP:** Leave enough funds after your action so your shops can be adequately restocked.
  **TIP:** Order quantity can be updated using the *modify* action.

There are three subtypes of shops:

- **Drink:** Sells beverages that quench guests' thirst.
- **Food:** Sells food that satisfies guests' hunger.
- **Specialty:** Provides unique services based on subclass:
  - Yellow (Souvenir Shops): Boosts guest happiness.
  - Blue (Info Booths): Informs guests about attractions and their prices.
  - Green (ATMs): Allows guests to withdraw additional funds.
  - Red (Billboards): Encourages guests to seek food and ATMs.

By default, guests do not actively seek out Specialty Shops; they visit them only if they pass by.

> **TIP:** Place specialty shops in high-traffic areas to increase the likelihood of guest interaction.

For exact parameters of each shop, see [All Shops](#).

---

## Staff

Staff are necessary for the smooth operation of your park, ensuring attractions run properly and guests remain satisfied. Multiple staff can occupy and work on the same tile at any given time. All staff have a daily salary, and some incur additional operating costs when performing tasks.

There are three subtypes of staff:

- **Janitors:** Move through the park toward dirty areas. When on dirty tiles, they clean them. Each cleaning action incurs an operating cost.
- **Mechanics:** Move toward rides that are broken down. When they reach a tile with a broken ride, they repair it. The time and operating cost of repairs depend on the ride's building cost and the mechanic(s) performing the repair.
- **Specialists:** Perform different roles based on subclass:
  - Yellow (Clowns): Increases the happiness of guests waiting in line.
  - Blue (Stockers): Restocks shops with low inventory.
  - Green (Park Criers): Informs guests about out-of-service or dirty attractions, as well as current line wait time for rides.
  - Red (Vendors): Provides food and drink to guests waiting in line.

For exact parameters of all staff members, see [All Employees](#).

---

## Subclasses & Research

Each ride, shop, and staff subtype has four subclasses, ordered by price: yellow (cheapest), blue, green, and red (most expensive). Generally, more expensive subclasses provide greater benefits. However, at times cheaper subclasses may better suit your needs.

In Easy mode, all attractions are available from the start and research is disabled. In Medium and Hard modes, you begin with only yellow subclasses and must perform research to unlock the rest. To do this, you must set your research speed (none/slow/medium/fast) and topics (ride/shop/staff subtypes). Research continues daily until you change your settings, run out of funds, or unlock all available subclasses for the chosen topics. If funds run out or all topics are complete, research speed automatically reverts to *none*. Progress is paused, not lost. Research performed provides a small amount of added value to the park in the form of intellectual property, but this value does not increase with research speed.

Research unlocks subclasses in the following order: blue → green → red. Once a subclass is unlocked, research continues to the next topic in your list.

> **TIP:** If for example, you wanted to unlock the red roller coaster as quickly as possible, set the research speed to *fast* and select only "roller coaster" as your research topic.

Research speed costs:

- **none:** $0/day — research halted.
- **slow:** $2000/day.
- **medium:** $8000/day.
- **fast:** $32000/day.

The default setting is a research speed of *none* and all attraction subtypes selected as topics.

---

## Guests

Guests are the heart of your park. The number of guests who visit depends on your park's rating and capacity.

**Capacity** determines both how many guests can be in the park at once and how many potential guests consider visiting. Capacity is determined entirely by the cumulative capacity of your rides.

> **TIP:** Since only rides increase capacity, a park with no rides will receive no visitors.

**Park Rating** determines the likelihood that potential guests decide to enter. New guests cannot enter if the park is already at capacity. Park rating increases with the total excitement of the rides and when guests leave the park happy. Park rating decreases when attractions are frequently out of service, the park is dirty, or the average ride intensity is too high or too low.

> **TIP:** Park ratings are calculated at the start of each day. In some cases, it may take a full day for the action you have taken to effect the park. In these cases, the park rating will only be reflected the subsequent day (two days after the action).

Each guest brings a certain amount of money. When they run out of funds or become angry, they leave. Each guest also has hunger, thirst, happiness, and energy levels. Hunger and thirst increase over time, while happiness slowly decreases. Hungry guests seek food shops, thirsty guests seek drink shops, and unhappy guests seek rides. If any of these needs become critical, the guest's happiness drops and they may leave the park.

> **TIP:** A guest's hunger and thirst will continue to increase as they wait in lines. Proximity to food and drink options is especially important for high capacity rides that have longer lines.

Guests favor novelty, preferring kinds of attractions they haven't visited before. They also favor nearby attractions but never visit the same attraction twice in a row. Visiting a ride they can't afford or that's out of service reduces their happiness.

Unhappy guests are more likely to litter, reducing cleanliness. Visiting dirty tiles or attractions decreases happiness further. If a ride is too dirty, guests may turn away.

In Hard mode, guests may have **preferences**, meaning they only enjoy certain attraction types. Visiting a ride that doesn't match their preference reduces their happiness. Providing guests with information through an Info Booth (blue Specialty Shop) allows them to identify attractions that suit them.

**TIP:** You can learn more about guests by surveying them using the *SurveyGuest* action. This reveals why guests left, their needs at departure, and more. You can survey up to 25 guests at a cost of $500 per guest.

### Park Value

The total value of the park is the sum of the total money, the money that would be made by selling all constructed attractions, and a flat amount per day of research (the value of the discovered intellectual property). Each day of slow, medium, and fast research adds $1500, $3000, and $6000 respectively to the park's value.

## All Rides

Note: Any existing ride can be sold for 66.0% of its building cost. A ride that breaks down incurs a cost equal to 4.5% of its building cost to repair.

### Carousels

#### Yellow

- Building Cost: 250
- Cost per Operation: 1
- Capacity: 6
- Max Ticket Price: 4
- Excitement: 1
- Intensity: 1
- Breakdown Rate: 0.001

#### Blue

- Building Cost: 1500
- Cost per Operation: 2
- Capacity: 14
- Max Ticket Price: 6
- Excitement: 4
- Intensity: 3
- Breakdown Rate: 0.002

#### Green

- Building Cost: 11500
- Cost per Operation: 30
- Capacity: 26
- Max Ticket Price: 14
- Excitement: 3
- Intensity: 4
- Breakdown Rate: 0.003

#### Red

- Building Cost: 24000
- Cost per Operation: 12
- Capacity: 24
- Max Ticket Price: 24
- Excitement: 8
- Intensity: 5
- Breakdown Rate: 0.005

### Ferris Wheels

#### Yellow

- Building Cost: 600
- Cost per Operation: 10
- Capacity: 10
- Max Ticket Price: 5
- Excitement: 2
- Intensity: 2
- Breakdown Rate: 0.006

#### Blue

- Building Cost: 7500
- Cost per Operation: 20
- Capacity: 20
- Max Ticket Price: 7
- Excitement: 5
- Intensity: 3

- Breakdown Rate: 0.009

**Green**

- Building Cost: 50000
- Cost per Operation: 55
- Capacity: 40
- Max Ticket Price: 15
- Excitement: 4
- Intensity: 6
- Breakdown Rate: 0.024

**Red**

- Building Cost: 75000
- Cost per Operation: 75
- Capacity: 30
- Max Ticket Price: 28
- Excitement: 9
- Intensity: 8
- Breakdown Rate: 0.032

### Roller Coasters

**Yellow**

- Building Cost: 1000
- Cost per Operation: 8
- Capacity: 4
- Max Ticket Price: 10
- Excitement: 3
- Intensity: 4
- Breakdown Rate: 0.01

**Blue**

- Building Cost: 18000
- Cost per Operation: 25
- Capacity: 12
- Max Ticket Price: 20
- Excitement: 7
- Intensity: 7
- Breakdown Rate: 0.02

**Green**

- Building Cost: 60000
- Cost per Operation: 45
- Capacity: 28
- Max Ticket Price: 34
- Excitement: 6
- Intensity: 9
- Breakdown Rate: 0.025

**Red**

- Building Cost: 100000
- Cost per Operation: 100
- Capacity: 22
- Max Ticket Price: 50
- Excitement: 10
- Intensity: 10
- Breakdown Rate: 0.04

## All Shops

Note: Any existing shop can be sold for 66.0% of its building cost.

### Drink Shops

**Yellow**

- Building Cost: 100
- Item Cost: 0
- Max Item Price: 3
- Thirst Reduction: 0.24

**Blue**

- Building Cost: 2250
- Item Cost: 1
- Max Item Price: 6
- Thirst Reduction: 0.6

**Green**

- Building Cost: 17500
- Item Cost: 3
- Max Item Price: 17
- Thirst Reduction: 0.96
- Happiness Boost: 0.4

*In addition to quenching thirst, green drink shops additionally provide a boost to guest happiness.*

**Red**

- Building Cost: 48000
- Item Cost: 5
- Max Item Price: 25
- Thirst Reduction: 0.72
- Energy Boost: 20
- Caffeinated Steps: 50

*Red drinks caffeinate guests, which boosts a guest's energy and allows them to move twice as fast*

## Food Shops

**Yellow**

- Building Cost: 200
- Item Cost: 1
- Max Item Price: 5
- Hunger Reduction: 0.2

**Blue**

- Building Cost: 3600
- Item Cost: 2
- Max Item Price: 9
- Hunger Reduction: 0.5

**Green**

- Building Cost: 28000
- Item Cost: 4
- Max Item Price: 18
- Hunger Reduction: 0.8
- Thirst Reduction: 0.4

*Green food shops both satiate hunger and quench thirst.*

**Red**

- Building Cost: 60000
- Item Cost: 8
- Max Item Price: 34
- Hunger Reduction: 1.0
- Happiness Boost: 0.6

*Red food shops sell luxury food. This greatly satiates hunger and increases happiness*

## Specialty Shops

**TIP:** Guests will not target specialty shops, and will only visit specialty shops if the walk adjacent to one.

**Yellow (Souvenir Shop)**

- Building Cost: 250
- Item Cost: 4
- Max Item Price: 15
- Happiness Boost: 0.5

*These provide a happiness boost to guests that buy them the first time, but this happiness boost diminishes with each subsequent souvenir purchased.*

**Blue (Information Booth)**

- Building Cost: 10000
- Item Cost: 2
- Max Item Price: 5

*These provide information to guests about the attractions in the park and ensures that guests only visit rides that fall within their budget and preferences.*

**Green (ATM)**

- Building Cost: 50000
- Item Cost: 3
- Max Item Price: 5
- Money Withdrawal: 64

*ATMs allow guests to withdraw more money. The amount of money withdrawn decreases exponentially with every subsequent withdrawal, to a minimum of 16*

**Red (Billboard)**

- Building Cost: 10000
- Item Cost: 0
- Max Item Price: 0
- Thirst Boost: 0.25
- Hunger Boost: 0.25

*Billboards make guests more hungry, thirsty, and happy. They will additionally reset the visit count of attractions (meaning that guests are more likely to revisit attractions), and, if the guest has less than $25, will direct the guest to an ATM.*

## All Employees

Employees perform up to 500 actions per day. For example, moving to an adjacent tile costs a move action.

### Janitors

Janitors clean a tile up to their cleaning threshold before moving to the next tile to clean. In addition to their salary, janitors incur $1 per cleaning action in cleaning supplies.

**Yellow**

- Daily Salary: 25
- Clean Rate: 0.028
- Cleaning Threshold: 0.85

**Blue**

- Daily Salary: 100
- Clean Rate: 0.075
- Cleaning Threshold: 0.95

*Blue, green, and red janitors move at double speed.*

**Green**

- Daily Salary: 500
- Clean Rate: 0.2
- Cleaning Threshold: 1.0

**Red**

- Daily Salary: 2000
- Clean Rate: 0.35
- Cleaning Threshold: 1.2

*Red janitors perform preventative cleaning, allowing them to clean a tile to 1.2; normally tiles have a maximum cleanliness of 1.0*

### Mechanics

**Yellow**

- Daily Salary: 15
- Repair Rate: 2

**Blue**

- Daily Salary: 100
- Repair Rate: 8

*Blue, green, and red mechanics move at double speed.*

**Green**

- Daily Salary: 250
- Repair Rate: 20

**Red**

- Daily Salary: 1000
- Repair Rate: 50

*Red mechanics also perform preventative maintenance -- this allows them to partially repair a ride before it breaks down, further reducing the time required for repairs.*

**Specialists**

**Yellow (Clown)**

- Daily Salary: 60
- Happiness Boost: 0.25

*Clowns move between different rides, increasing the happiness of guests waiting in line.*

**Blue (Stocker)**

- Daily Salary: 350
- Stocking Rate: 0.1
- Max Inventory: 100
- Idle Ticks: 30
- Restock Threshold: 0.25

*Stockers restock the inventory of shops when the proportion of the shop's remaining inventory drops below the Restock Threshold. When this happens a stocker will move to an entrance or exit, purchase 10.0% of the daily order quantity, and take it to the shop. Stockers can carry at most 100 units of inventory and will not order more than this quantity. Stockers will stop making new purchases in the last 30 actions of the day to prevent them from restocking a shop immediately before closing.*

**Green (Crier)**

- Daily Salary: 250

*Criers move between attractions, providing guests about information related to the cleanliness and operation (i.e., if an attraction is out of service) of attractions, as well as current line wait times. This prevents guests from visiting out of service attractions, and makes them more likely to visit cleaner attractions and those with shorter wait times.*

**Red (Vendor)**

- Daily Salary: 300
- Hunger Reduction: 0.3
- Thirst Reduction: 0.4

*Vendors move between rides, providing food and drink to guests waiting in line. Vendors do not incur additional costs or make additional profits from their activities.*

## Action Space

Actions must be written as python function calls with keyword arguments. These action functions must be in the following format:

```
action_name(param_1=<param1_value>, param_2=<param2_value>, ... )
```

The full list of available actions, including the action names, parameters, and a description of what they do is below:

---

**place**
*Description*: Place an entity (ride or shop or staff) in the amusement park
*Parameters*:

- x: The x position of the entity
- y: The y position of the entity
- type: The type of entity. Must be one of ride or shop or staff
- subtype: The subtype of the entity to be placed. For rides, must be one of carousel, ferris wheel, or roller coaster. For shops, must be one of drink, food, or specialty. For staff, must be one of janitor, mechanic, or specialist
- subclass: The specific instance of the entity to be placed. Must be one of yellow, blue, green, or red
- price: The price of the ticket or item. Only for rides and shops.
- order_quantity: The maximum order_quantity of inventory. Only for shops.

---

**move**
*Description*: Move an entity (ride or shop or staff) in the amusement park
*Parameters*:

- type: The type of entity. Must be one of ride or shop or staff
- subtype: The subtype of the entity to be moved. For rides, must be one of carousel, ferris wheel, or roller coaster. For shops, must be one of drink, food, or specialty. For staff, must be one of janitor, mechanic, or specialist
- subclass: The specific instance of the entity to be moved. Must be one of yellow, blue, green, or red
- x: The current x position of the entity
- y: The current y position of the entity
- new_x: The new x position
- new_y: The new y position

---

**remove**
*Description*: Remove an entity (ride or shop or staff). If the entity is a ride or shop, it will be sold for a price.
*Parameters*:

- type: The type of entity. Must be one of ride or shop or staff
- subtype: The subtype of the entity to be removed. For rides, must be one of carousel, ferris wheel, or roller coaster. For shops, must be one of drink, food, or specialty. For staff, must be one of janitor, mechanic, or specialist

- subclass: The specific instance of the entity to be removed. Must be one of yellow, blue, green, or red
- x: The x position of the entity
- y: The y position of the entity

---

**modify**
*Description*: Change the price and (as applicable) order_quantity of inventory for an attraction
*Parameters*:

- type: The type of attraction. Must be one of ride or shop
- x: The x position of the attraction
- y: The y position of the attraction
- price: The new price
- order_quantity: The new maximum order_quantity of inventory. Only for shops.

---

**set_research**
*Description*: Set the research speed and topic(s) for the amusement park. Only available in medium or hard difficulty.
*Parameters*:

- research_speed: The research speed. Must be one of none, slow, medium, or fast
- research_topics: The topics(s) of research. Must be a subset of ['carousel', 'ferris_wheel', 'roller_coaster', 'drink', 'food', 'specialty']

---

**survey_guests**
*Description*: Retrieve a sample of information from guests
*Parameters*:

- num_guests: The number of guests to survey

---

**add_path**
*Description*: Add a path tile to the amusement park. Only available in hard difficulty.
*Parameters*:

- x: The x position of the path tile
- y: The y position of the path tile

---

**remove_path**
*Description*: Remove a path tile from the amusement park. Only available in hard difficulty.
*Parameters*:

- x: The x position of the path tile
- y: The y position of the path tile

---

**add_water**
*Description*: Add a water tile to the amusement park. Only available in hard difficulty.
*Parameters*:

- x: The x position of the water tile
- y: The y position of the water tile

---

**remove_water**
*Description*: Remove a water tile from the amusement park. Only available in hard difficulty.
*Parameters*:

- x: The x position of the water tile
- y: The y position of the water tile

---

**wait**
*Description*: runs the day without any new action
*Parameters*:
No parameters

---

# Sandbox Mode

The game can be placed in sandbox mode to help with learning or exploring the environment. In sandbox mode, players have access to new actions that are useful for experimenting with the game. These actions are likewise written as python function calls (see Action Space).

When in sandbox mode, the game permits the player to perform 100 normal actions and an unlimited number of sandbox actions.

## Sandbox Action Space

The full list of available sandbox actions, including the action names, parameters, and a description of what they do is below:

---

**undo_day**
*Description*: Undo the last day, reverting to the previous state.
*Parameters*:

No parameters

---

**max_money**
*Description*: Set the park's money to the maximum value ($99,999,999).
*Parameters*:
No parameters

---

**max_research**
*Description*: Unlock all attractions and staff.
*Parameters*:
No parameters

---

**reset**
*Description*: Reset the park to initial state
*Parameters*:
No parameters

---

**change_settings**
*Description*: Change the park's difficulty and layout settings, and reset the park using the new settings.
*Parameters*:

- difficulty: The difficulty level. Must be one of easy or medium
- layout: The park layout. Must be one of ['diagonal_squares', 'the_ladder', 'two_paths', 'the_line', 'the_fork'].

---



\newpage
\section*{NeurIPS Paper Checklist}

\begin{enumerate}

\item {\bf Claims}
    \item[] Answer: \answerYes{} %
    \item[] Justification: The abstract describes the \envname\ environment and benchmark, and its purpose of evaluating long-horizon planning, sample efficient learnin, spatial reasoning, under uncertainty. The paper contains sections and experiments corresponding to these points.

\item {\bf Limitations}
    \item[] Question: Does the paper discuss the limitations of the work performed by the authors?
    \item[] Answer: \answerYes{} %
    \item[] Justification: See the Limitations, Appendix~\ref{sec:limitations}.

\item {\bf Theory assumptions and proofs}
    \item[] Question: For each theoretical result, does the paper provide the full set of assumptions and a complete (and correct) proof?
    \item[] Answer: \answerNA{} %
    \item[] Justification: This paper provides a new benchmark and empirical findings.

    \item {\bf Experimental result reproducibility}
    \item[] Question: Does the paper fully disclose all the information needed to reproduce the main experimental results of the paper to the extent that it affects the main claims and/or conclusions of the paper (regardless of whether the code and data are provided or not)?
    \item[] Answer: \answerYes{} %
    \item[] Justification: See Sections~\ref{app:implementation_details} and~\ref{sec:method}. Furthermore, we release our code with detailed instructions on how to set up and evaluate and agent on \envname.

\item {\bf Open access to data and code}
    \item[] Question: Does the paper provide open access to the data and code, with sufficient instructions to faithfully reproduce the main experimental results, as described in supplemental material?
    \item[] Answer: \answerYes{} %
    \item[] Justification: We publicly release our benchmark, including the code for both the environment and our baselines. See supplemental, a GitHub repository will be released on publication.

\item {\bf Experimental setting/details}
    \item[] Question: Does the paper specify all the training and test details (e.g., data splits, hyperparameters, how they were chosen, type of optimizer) necessary to understand the results?
    \item[] Answer: \answerYes{} %
    \item[] Justification: See Appendix~\ref{app:implementation_details}, as well as prompts in Appendix~\ref{app:prompts}. Furthermore, we release our code.

\item {\bf Experiment statistical significance}
    \item[] Question: Does the paper report error bars suitably and correctly defined or other appropriate information about the statistical significance of the experiments?
    \item[] Answer: \answerYes{} %
    \item[] Justification: We repeat our experiments over 3 different random seeds, reporting the mean and standard deviation.

\item {\bf Experiments compute resources}
    \item[] Question: For each experiment, does the paper provide sufficient information on the computer resources (type of compute workers, memory, time of execution) needed to reproduce the experiments?
    \item[] Answer: \answerYes{} %
    \item[] Justification: Yes, see App~\ref{app:cost}
    
\item {\bf Code of ethics}
    \item[] Question: Does the research conducted in the paper conform, in every respect, with the NeurIPS Code of Ethics \url{https://neurips.cc/public/EthicsGuidelines}?
    \item[] Answer: \answerYes{} %
    \item[] Justification: \envname\ itself contains no information from human participants. Crowdsourcing/contract work was not used, instead the leaderboard is open to the general public. The relevant ethics review process was undertaken, the leaderboard collects no personally identifiable information, and it contains a EULA (see Appendix~\ref{app:eula}). We do not believe that \envname\ poses any risks of potential harms, beyond the very broad consequence that providing a benchmark for improved long-horizon reasoning encourages the development of AI systems that are economically useful.  However it would be this followup work, not the benchmark itself, that poses a risk.

\item {\bf Broader impacts}
    \item[] Question: Does the paper discuss both potential positive societal impacts and negative societal impacts of the work performed?
    \item[] Answer: \answerNA{} %
    \item[] Justification: See the answer to the previous section. In short, a benchmark is not directly deployable to do harm, and the benchmark itself does not contain personal information or target harmful skills.
    
\item {\bf Safeguards}
    \item[] Question: Does the paper describe safeguards that have been put in place for responsible release of data or models that have a high risk for misuse (e.g., pre-trained language models, image generators, or scraped datasets)?
    \item[] Answer: \answerNA{} %
    \item[] Justification: This paper does not release a model.

\item {\bf Licenses for existing assets}
    \item[] Question: Are the creators or original owners of assets (e.g., code, data, models), used in the paper, properly credited and are the license and terms of use explicitly mentioned and properly respected?
    \item[] Answer: \answerYes{} %
    \item[] Justification: The papers for the relevant LLMs used in our experiments are cited (see Section~\ref{sec:open_end}). Otherwise, \envname\ contains assets created by the authors.

\item {\bf New assets}
    \item[] Question: Are new assets introduced in the paper well documented and is the documentation provided alongside the assets?
    \item[] Answer: \answerYes{} %
    \item[] Justification: The environment is described thoroughly. See Section~\ref{sec:method}, as well as the full game manual in Appendix~\ref{app:documentation}. The environment is released under an MIT License (see code).

\item {\bf Crowdsourcing and research with human subjects}
    \item[] Question: For crowdsourcing experiments and research with human subjects, does the paper include the full text of instructions given to participants and screenshots, if applicable, as well as details about compensation (if any)? 
    \item[] Answer: \answerYes{} %
    \item[] Justification: We do not perform a human study. Instead, we provide a leaderboard that is open to the public. No crowdsourcing or contract work was used. A screenshot of the interface is provided in Figure~\ref{fig:main_fig}, and the full game manual is available at Appendix~\ref{app:documentation}. See Appendix~\ref{app:gui}.

\item {\bf Institutional review board (IRB) approvals or equivalent for research with human subjects}
    \item[] Question: Does the paper describe potential risks incurred by study participants, whether such risks were disclosed to the subjects, and whether Institutional Review Board (IRB) approvals (or an equivalent approval/review based on the requirements of your country or institution) were obtained?
    \item[] Answer: \answerYes{} %
    \item[] Justification: There are no risks to participants, and a review was performed (see App~\ref{app:gui}). Further information will be added after anonymity requirements are lifted.

\item {\bf Declaration of LLM usage}
    \item[] Question: Does the paper describe the usage of LLMs if it is an important, original, or non-standard component of the core methods in this research? Note that if the LLM is used only for writing, editing, or formatting purposes and does \emph{not} impact the core methodology, scientific rigor, or originality of the research, declaration is not required.
    \item[] Answer: \answerNA{} %
    \item[] Justification: Beyond their use in the agents evaluated (documented in Section~\ref{sec:open_end}), LLMs were only used for text editing and basic code assistance.

\end{enumerate}

\end{document}